\pdfoutput=1
\documentclass[10pt,twocolumn,letterpaper]{article}

\usepackage[pagenumbers]{cvpr} 

\usepackage{graphicx}
\usepackage{amsmath}
\usepackage{amssymb}
\usepackage{booktabs}
\usepackage{array}
\usepackage{microtype}
\usepackage{capt-of}
\usepackage[breakable]{tcolorbox}

\newtcolorbox{promptbox}{breakable, colback=gray!6, colframe=gray!45,
  boxrule=0.4pt, arc=1pt, left=4pt, right=4pt, top=3pt, bottom=3pt,
  before skip=6pt, after skip=6pt, fontupper=\footnotesize}

\renewcommand{\paragraph}[1]{\vspace{.35em}\noindent\textbf{#1.}}
\definecolor{cvprblue}{rgb}{0.21,0.49,0.74}
\usepackage[pagebackref,breaklinks,colorlinks,allcolors=cvprblue]{hyperref}

\title{PROVE: Predict-to-Specify and Observe-to-Verify for Human-Aligned Evaluation of Mobile Manipulation}

\author{Bowen Lyu \quad Yunan Liu \quad Liaoyuan Fan \quad Qinchuan Cheng\\
Jiafei Duan \quad Cheston Tan \quad Shijie Li
}

\begin{document}
\maketitle

\begin{abstract}
Judging open-world robot manipulation is hard because layouts, articulated fixtures, and valid goal configurations vary widely.
On RoboCasa365, simulator predicates agree with human judgments on only 69.09\% of rollouts, and a frontier VLM judge on 86.12\%.
We present \textbf{PROVE} (Predict-to-Specify, Observe-to-Verify), which first specifies what success should look like and then checks the outcome.
PROVE refines success conditions from the instruction,
the initial scene, a target-object reference image,
and a generated goal image.
The goal image only sharpens the conditions and never serves as evidence.
On 6,500 human-annotated rollouts over 65 atomic actions, PROVE reaches \textbf{96.83\%} agreement.
Applied to open-source policies, it reports far lower success than native checks and changes their ranking.
To reduce evaluation overhead in policy learning, we distill PROVE into \textbf{PROVE-Lite}, a 4B evaluator that needs no reference images.
In one forward pass, PROVE-Lite predicts the outcome and the terminal state of each task condition.
It reaches \textbf{91.20\%} agreement while running about 38$\times$ faster than PROVE.
Mapped to a severity-aware reward, its diagnoses improve PPO post-training over both simulator and binary rewards.
Project page: \url{https://sj-li.com/projects/prove/}.
\end{abstract}

\section{Introduction}
\label{sec:intro}

\begin{figure}[!t]
\centering
\scriptsize
\setlength{\tabcolsep}{2pt}
\renewcommand{\arraystretch}{1.1}

\setbox0=\hbox{%
  \begin{tabular}{@{}c@{\hspace{2pt}}c@{}}
  \rotatebox{90}{\makebox[0.7\dimexpr0.265\columnwidth\relax][c]{\textbf{Before}}} &
  \includegraphics[width=0.7\dimexpr0.265\columnwidth\relax]
  {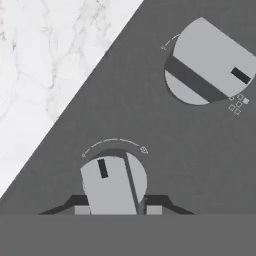} \\[1pt]
  \rotatebox{90}{\makebox[0.7\dimexpr0.265\columnwidth\relax][c]{\textbf{After}}} &
  \includegraphics[width=0.7\dimexpr0.265\columnwidth\relax]
  {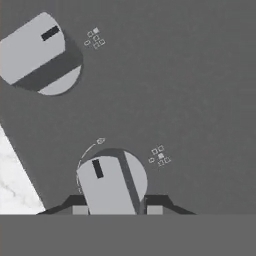}
  \end{tabular}%
}
\dimen0=\dimexpr\ht0+\dp0\relax

\long\def\imagecell#1{%
  \begin{minipage}[t]{\linewidth}
  \vspace{0pt}%
  \parbox[c][\dimen0][c]{\linewidth}{\centering #1\par}%
  \end{minipage}%
}

\begin{tabular}{
@{}
>{\raggedright\arraybackslash}p{0.13\columnwidth}
>{\raggedright\arraybackslash}p{0.265\columnwidth}
>{\raggedright\arraybackslash}p{0.265\columnwidth}
>{\raggedright\arraybackslash}p{0.265\columnwidth}
@{}
}
\toprule
& \textbf{(a) CoffeeSetupMug} & \textbf{(b) LowerHeat} & \textbf{(c) CloseBlenderLid} \\
\midrule
Instruction &
Place the mug under the coffee dispenser. &
Lower the heat of the kettle. &
Close the blender lid securely. \\
\midrule
\imagecell{\raggedright Images} &
\imagecell{\includegraphics[width=\linewidth]{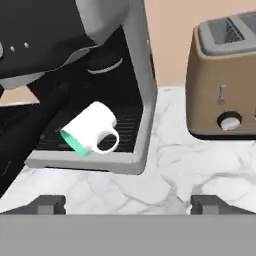}} &
\imagecell{\usebox0} &
\imagecell{\includegraphics[width=\linewidth]{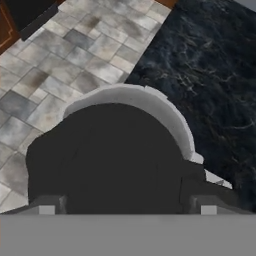}} \\
\midrule
RoboCasa &
\textbf{SUCCESS}\newline Mug near dispenser; gripper away. &
\textbf{SUCCESS}\newline Knob angle within the low-heat range. &
\textbf{SUCCESS}\newline Lid touches the blender; gripper away. \\
\midrule
PROVE &
\textbf{FAILURE}\newline Mug lies on its side, opening sideways. &
\textbf{FAILURE}\newline Before vs.\ after: heat was raised. &
\textbf{FAILURE}\newline Lid does not fully cover the opening. \\
\midrule
Human &
\textbf{FAILURE}\newline Mug lies on its side. &
\textbf{FAILURE}\newline Knob turned up, still in the low range. &
\textbf{FAILURE}\newline Lid does not cover the opening. \\
\bottomrule
\end{tabular}

\caption{\textbf{Native checks accept failures that humans reject.}
RoboCasa passes all three because each check tests only part of the task.
In (b), only a before/after comparison shows that the knob was turned up.}
\label{fig:native_false_success}
\end{figure}

Mobile manipulation benchmarks decompose household tasks into atomic actions such as grasping, placing, closing doors, and turning knobs.
Judging these actions reliably matters for benchmark credibility \cite{robocasa2024,robocasa365_2026,zhou2025autoeval} and for policy improvement \cite{zang2025rlinfvla}.
In open-world scenes, this is harder than it looks.
Layouts and articulated fixtures vary across kitchens, and one instruction often admits several valid goal configurations.
An evaluator should tell whether the robot succeeded and which physical conditions remain unmet.
Ideally, this diagnosis should also help improve the policy.

Existing evaluators fall short.
Simulation benchmarks check fixed predicates over privileged object states.
These states are unavailable on physical robots, and the predicates cannot anticipate every valid outcome.
On RoboCasa365 \cite{robocasa365_2026}, they agree with human judgments on only \textbf{69.09\%} of rollouts.
They err in both directions, accepting failures and rejecting successes, and never explain why (\cref{fig:native_false_success}).
VLM judges \cite{du2023successvqa,sermanet2024robovqa} avoid privileged state, but instruction-and-terminal-only judging omits the initial scene context.
Prompted with the instruction alone, they turn plausible yet irrelevant expectations into requirements.
For example, they may demand closing a cabinet door that was already open.
Such a judge reaches \textbf{82.52\%} agreement with our 8B backbone and \textbf{86.12\%} with a frontier model, GPT-6.1 Sol~\cite{openai2026gpt61sol}.

Human judges do better because they separate two questions.
What should success look like in this scene, and did it actually happen?
They first set the standard from the task, the scene, and the expected outcome, and only then consult the evidence.
VLMs understand instructions, objects, and spatial relations \cite{qwen3vl2025}.
Image-editing models can predict the intended post-action configuration \cite{brooks2023instructpix2pix,black2024susie}.
Generated images, however, contain artifacts and implausible contact.
They tell us about the goal but are unreliable as a reference for matching.
We therefore let predictions specify and observations verify.
The goal image only sharpens the success conditions.
The terminal observations alone decide whether those conditions hold, so the outcome never shapes its own standard.

\begin{figure*}[t]
  \centering
  \includegraphics[width=\textwidth]{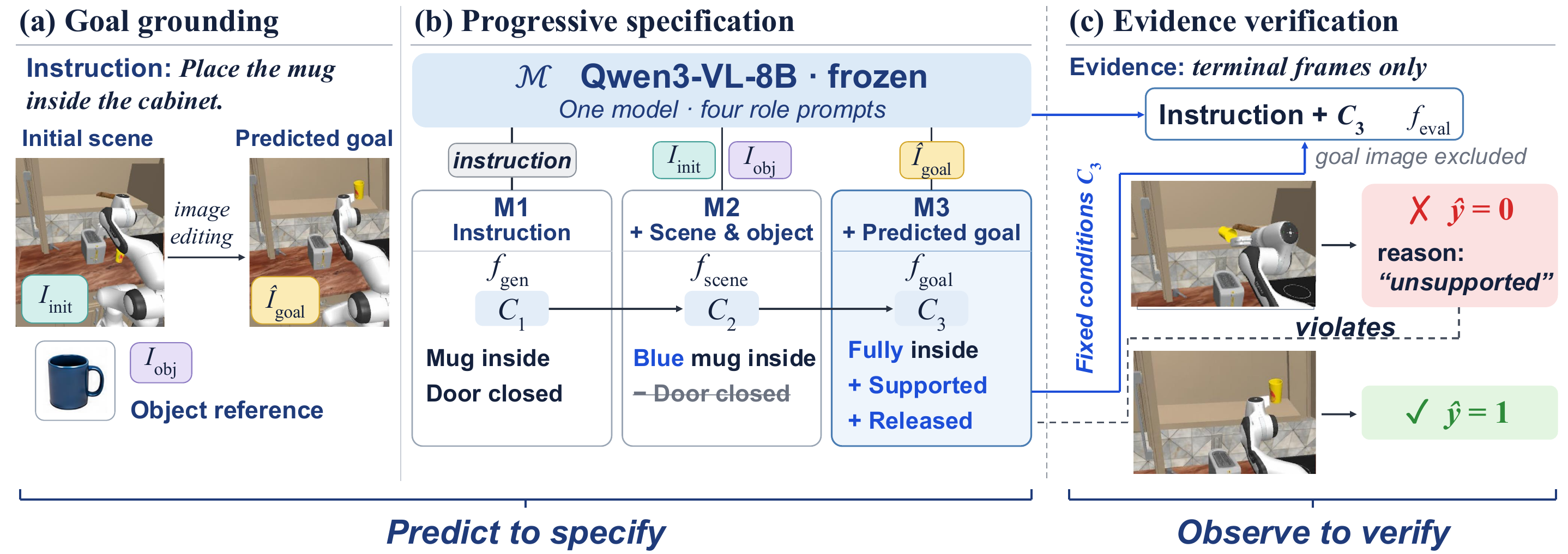}
  \caption{\textbf{Overview of PROVE.}
  (a) Inputs: the instruction, the initial scene $I_\mathrm{init}$, an object reference $I_\mathrm{obj}$,
  and a goal image $\hat I_\mathrm{goal}$ predicted by image editing.
  (b) One frozen MLLM $\mathcal{M}$ refines the success conditions under three role prompts.
  Highlighted words are refined, ``+'' marks added conditions, and struck-out conditions are removed.
  (c) With $f_\mathrm{eval}$, $\mathcal{M}$ checks the fixed conditions $C_3$ against terminal frames only.
  It outputs a label $\hat y$ and a reason; the goal image is never used here.}
  \label{fig:prove}
\end{figure*}

Following this principle, we build \textbf{PROVE}, an evaluation harness that mirrors the human process (\cref{fig:prove}).
One frozen MLLM refines success conditions in three stages.
It starts from the instruction alone (\textbf{M1}) and adds the initial scene and target object (\textbf{M2}).
The last stage adds a generated goal image (\textbf{M3}).
On 6,500 human-annotated rollouts over 65 atomic actions, this progressive grounding raises agreement to \textbf{96.83\%}.
Ablations confirm that the goal image helps only as specification, and that the gain transfers across backbones.
Applied to four open-source policies, PROVE reports far lower success than native checks and reorders them.
When the two disagree, blind human review sides with PROVE in \textbf{95.42\%} of cases.

\begin{figure}[t]
  \centering
  \includegraphics[width=\linewidth]{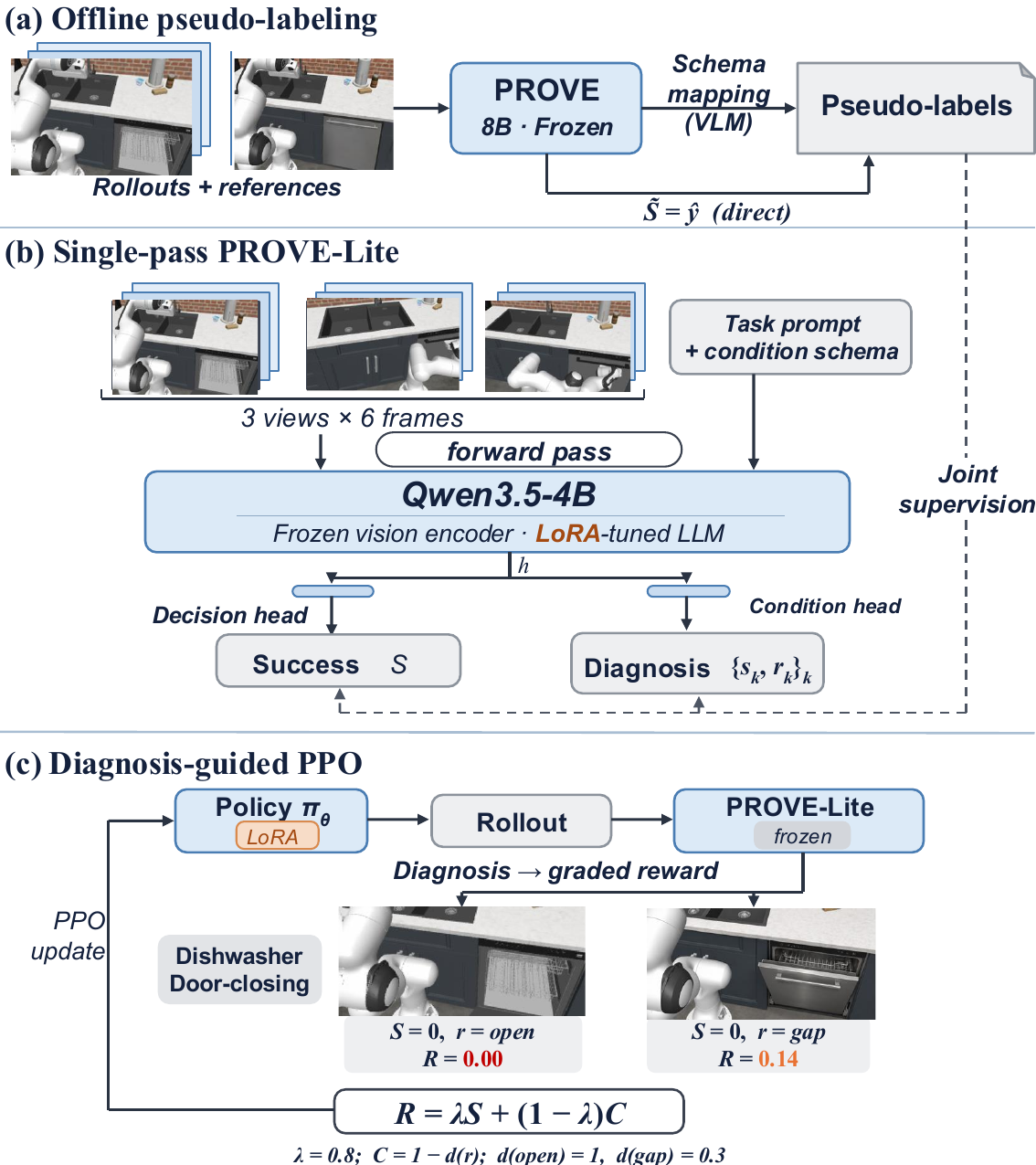}
\caption{\textbf{Overview of PROVE-Lite.}
  (a) PROVE labels training rollouts offline; a VLM maps its conditions and verdict to the condition schema.
  (b) PROVE-Lite reads three views of six terminal frames in one forward pass.
  Two light heads share the hidden state $\mathbf{h}$.
  They predict success $S$ and a per-condition diagnosis $\{s_k, r_k\}_k$.
  (c) The frozen PROVE-Lite turns the diagnosis into a graded reward
  $R = \lambda S + (1-\lambda)C$ with $C = 1 - \sum_k w_k\, d(r_k)$, which drives PPO.
  Examples show a door-closing task.}
  \label{fig:prove_lite}
\end{figure}

PROVE suits offline evaluation, where each rollout is judged once.
Policy training is different: thousands of rollouts must be judged on the fly.
There, PROVE's multi-stage inference and reference assets become too costly.
We therefore distill \cite{hinton2015distilling} it into \textbf{PROVE-Lite}, an evaluator that needs no reference images (\cref{fig:prove_lite}).
In a single forward pass, PROVE-Lite predicts the outcome and the terminal state of each task condition.
Trained only on PROVE's pseudo-labels, it reaches \textbf{91.20\%} agreement while running about 38$\times$ faster.

We validate PROVE-Lite where it is meant to be used: as the reward in policy post-training.
Its diagnoses describe how far each condition is from its target.
They can therefore grade failures instead of treating them alike.
A narrow door gap, for instance, is penalized less than a wide-open door.
We map each diagnosed state to a fixed severity.
The resulting terminal reward gives failed rollouts bounded partial credit while keeping binary success dominant.
In controlled PPO \cite{schulman2017ppo} post-training, it beats simulator rewards by \textbf{4.58} points and binary PROVE-Lite rewards by \textbf{2.00} points.

Our contributions are three-fold:
\begin{itemize}
    \item \textbf{Predict-to-Specify, Observe-to-Verify evaluation.} PROVE grounds success conditions in the task, the scene, and a generated goal image. It verifies them against real observations only.
    \item \textbf{Reference-free evaluation for policy learning.} PROVE-Lite is distilled from PROVE without manual labels. Beyond the outcome, it diagnoses the terminal state of each task condition in one forward pass.
    \item \textbf{Diagnosis-guided terminal reward.} The reward grades failed rollouts by the physical severity of their unmet conditions. It outperforms simulator and binary rewards in PPO post-training.
\end{itemize}

\section{Related Work}
\label{sec:related}

\paragraph{Evaluating Robot Manipulation}
Simulation benchmarks including RoboCasa~\cite{robocasa2024,robocasa365_2026}, RLBench~\cite{james2020rlbench}, and LIBERO~\cite{liu2023libero} use simulator-state checks. BEHAVIOR~\cite{srivastava2022behavior,li2023behavior1k} formalizes initial and goal conditions with predicates. For physical robots, AutoEval~\cite{zhou2025autoeval} learns success classifiers. Following the broader model-as-judge paradigm~\cite{zheng2023judging}, SuccessVQA~\cite{du2023successvqa} and RoboVQA~\cite{sermanet2024robovqa} formulate judgments as visual questions. RoboJev~\cite{chen2026robojev} adapts the JEV-27B-VL judge~\cite{autotrust2026jev27bvl} with LoRA and reads success from two calibrated tokens in one forward pass. Eval-Actions~\cite{liu2026evalactions} pairs policy rollouts, failures included, with teleoperation data to grade execution quality. PROVE grounds evaluation criteria in scene-specific visual context and fixes them before inspecting terminal evidence.

\paragraph{Generated Goals in Robot Learning}
SuSIE~\cite{black2024susie} and UniPi~\cite{du2023unipi} generate goal images or video plans for control; Gen2Act~\cite{bharadhwaj2024gen2act} conditions robot policies on generated human videos. VoxPoser~\cite{huang2023voxposer} grounds language-derived affordances and constraints in 3D value maps for planning. PROVE uses a generated goal to refine evaluation criteria instead. Since generated images can contain artifacts and cannot enumerate all valid outcomes, we retain textual conditions and withhold the generated image during verification.

\paragraph{Rewards from Foundation Models}
VLM rewards use image--text similarity~\cite{rocamonde2024vlmrm,baumli2023vlmrewards}, video--language matching~\cite{sontakke2023roboclip}, and preference feedback~\cite{wang2024rlvlmf}; Eureka~\cite{ma2024eureka} instead uses an LLM to write reward code. LIV~\cite{ma2023liv} jointly learns representations and value-based rewards. GVL~\cite{ma2025gvl}, Robometer~\cite{liang2026robometer}, and ProcVLM~\cite{procvlm2026} estimate progress using in-context reasoning, trajectory comparisons, and procedural structure, respectively. Closest to our terminal grading, RoboReward~\cite{lee2026roboreward} already scores partial and near-complete outcomes under an end-of-episode rubric. Our distinction is condition-level terminal diagnosis with a fixed class-to-severity mapping, learned from a teacher that separates criterion specification from verification.

\paragraph{Robot Failure Explanation}
REFLECT~\cite{liu2023reflect} summarizes execution experience for failure explanation and correction. AHA~\cite{duan2025aha} provides failure feedback for planning and reward refinement; Guardian~\cite{pacaud2026guardian} uses structured reasoning to detect manipulation failures. PROVE-Lite restricts diagnosis to terminal evidence: it predicts a closed set of per-condition physical states mapped directly to reward severity, without inferring an unobserved execution history.

\section{Evidence-Grounded Action Evaluation}
\label{sec:method}

\subsection{Evaluation Formulation}
\label{sec:method_formulation}

Simulator predicates \cite{robocasa2024,robocasa365_2026} cannot capture the diverse layouts, articulated fixtures, and multiple valid goal configurations of open-world manipulation. Human judges instead combine task understanding, scene context, expected outcomes, and visual evidence. We abstract this process into three steps: (1) \textbf{task and scene grounding}, which establishes the intended physical outcomes in the current environment; (2) \textbf{goal anticipation}, which predicts the expected post-action configuration; and (3) \textbf{evidence-grounded verification and diagnosis}, which checks each condition against actual observations and identifies the primary deficiency when execution fails.

We instantiate this formulation on the 65 atomic mobile manipulation actions in RoboCasa365 \cite{robocasa365_2026}, where each evaluation instance corresponds to one rollout specified by a natural-language instruction $\mathcal{I}$. Our MLLM-based evaluation framework, \textbf{PROVE}, strictly decouples goal specification from outcome verification. Before inspecting the terminal observation $I_{\text{final}}$, PROVE first states the required physical postconditions from $\mathcal{I}$ and progressively refines them using context that describes the goal rather than the execution outcome: the initial scene $I_{\text{init}}$, a target-object reference $I_{\text{obj}}$, and a predicted goal image $\hat{I}_{\text{goal}}$ (\Cref{sec:method_teacher}). Only after locking in these criteria does PROVE examine $I_{\text{final}}$ to output a binary decision $\hat{y} \in \{0,1\}$ alongside a primary failure reason $\rho$, ensuring that the actual outcome never shapes the standard by which it is judged (\Cref{sec:method_verification}).

\subsection{Progressive Condition Grounding}
\label{sec:method_teacher}

Across condition grounding and outcome verification, PROVE uses a single frozen MLLM $\mathcal{M}$ \cite{qwen3vl2025} instantiated with role-specific prompts: $f_{\text{gen}}$ for initial condition generation, $f_{\text{scene}}$ and $f_{\text{goal}}$ for scene- and goal-guided refinement, and $f_{\text{eval}}$ for terminal verification and failure diagnosis.

Task instructions specify the intended action but often leave scene-dependent completion criteria implicit. For example, an instruction to place an object inside a cabinet does not fully specify the target region, acceptable spatial tolerance, or required support condition. The expected physical outcomes must therefore be grounded in the actual scene before execution can be evaluated.

We progressively refine textual verification conditions through three stages (\cref{fig:prove}b). Instruction-derived semantic conditions establish the initial specification (\textbf{M1}). Scene context and object references resolve ambiguities in that specification (\textbf{M2}). Finally, a predicted goal image provides additional guidance about the intended post-action configuration (\textbf{M3}). The resulting conditions form the explicit criteria for subsequent verification.

\paragraph{Instruction-Derived Semantic Conditions (M1)}
The first stage establishes task-level expectations using semantic and physical priors. Given the task instruction $\mathcal{I}$, the MLLM $\mathcal{M}$ derives a set of concise physical postconditions:
\begin{equation}
    C_1 = f_{\text{gen}}(\mathcal{I}).
\end{equation}
Without scene context, M1 relies on generic language priors and may turn plausible but task-irrelevant expectations into required postconditions. For example, an instruction to place an object inside a cabinet may lead M1 to require closing a cabinet door that was already open, or moving the robot arm out of the camera view. Neither requirement necessarily follows from the instruction.

\paragraph{Scene-Grounded Refinement (M2)}
The second stage grounds these expectations in the actual environment. It refines $C_1$ using the initial scene observation $I_{\text{init}}$ and the target-object reference $I_{\text{obj}}$:
\begin{equation}
    C_2 = f_{\text{scene}}(C_1, \mathcal{I},
    I_{\text{init}}, I_{\text{obj}}).
    \label{eq:m2}
\end{equation}
The initial scene observation and object reference help identify the target object (e.g., distinguishing the target mug from visually similar containers) and establish which fixtures are already in the desired state. This prevents the evaluator from imposing unnecessary changes to the initial scene. However, because these inputs describe only the pre-action scene, they cannot fully specify the intended post-action geometry or spatial tolerances.

\paragraph{Goal-Guided Condition Refinement (M3)}
The third stage anticipates the expected post-action configuration using an image-editing generative model \cite{brooks2023instructpix2pix,black2024susie}. Its candidate goal image $\hat{I}_{\text{goal}}$ provides additional spatial and physical guidance for refining the verification conditions:
\begin{equation}
  C_3 = f_{\text{goal}}(C_2, \mathcal{I},
    I_{\text{init}}, I_{\text{obj}}, \hat{I}_{\text{goal}}).
    \label{eq:m3}
\end{equation}

Unlike generic semantic priors or pre-action observations, $\hat{I}_{\text{goal}}$ provides a concrete visual hypothesis of the post-action arrangement, enabling $\mathcal{M}$ to clarify target spatial relationships, placement regions, and support configurations. Crucially, $\mathcal{M}$ distills only high-level geometric and relational cues from $\hat{I}_{\text{goal}}$ into the explicit textual criteria $C_3$, abstracting away low-level appearance variations. Because $\hat{I}_{\text{goal}}$ is used strictly during condition specification and withheld during outcome verification, it helps define the expected goal standard without ever serving as evidence of task completion.

\subsection{Evidence-Grounded Outcome Verification}
\label{sec:method_verification}

Once the conditions $C_3$ are fixed, $\mathcal{M}$ acts as an evidence verifier, checking each condition against the terminal observation $I_{\text{final}}$ to output a binary outcome $\hat{y} \in \{0,1\}$ along with a primary failure reason $\rho$:
\begin{equation}
    (\hat{y}, \rho) = f_{\text{eval}}(
    \mathcal{I}, C_3, I_{\text{final}}).
    \label{eq:eval_teacher}
\end{equation}
Specifically, $f_{\text{eval}}$ inspects all three camera views across six terminal frames (18 images in total).
Every postcondition in $C_3$ must be positively confirmed by the terminal evidence; partial cues are not enough.
Multiple views keep a nearby gripper that occludes a placed object from being misjudged as an incomplete release.
Later frames capture transient events such as object settling or appliance activation.
When an execution fails ($\hat{y}=0$), $\rho$ records the primary observable deficiency in the terminal state, without speculating about earlier events.
The pair $(\hat{y}, \rho)$ later supervises PROVE-Lite (\Cref{sec:method_jev}).

\section{Efficient Evaluation and Policy Learning}
\label{sec:jev_rl}

PROVE suits offline evaluation, where each rollout is judged once and accuracy outweighs cost. An evaluator is most valuable, however, as a reward signal: online post-training \cite{zang2025rlinfvla} must judge thousands of fresh rollouts inside the training loop, which PROVE's reference assets, goal-image synthesis, and multi-stage inference make too costly. We therefore distill \cite{hinton2015distilling} it into \textbf{PROVE-Lite}, a 4B evaluator that needs no reference images and reads out the binary outcome and the terminal state of each task condition from a single hidden state in one forward pass (\Cref{sec:method_jev}), and convert this diagnosis into a terminal reward for PPO \cite{schulman2017ppo} (\Cref{sec:method_rl}).

\subsection{PROVE-Lite: Reference-Free Evaluator Distillation}
\label{sec:method_jev}

\paragraph{Single-Pass Readout}
We call \textbf{PROVE-Lite} reference-free because it needs no initial or generated reference images: no $I_{\text{init}}$, no $I_{\text{obj}}$, and no $\hat{I}_{\text{goal}}$. Its visual input is the terminal observation $I_{\text{final}}$: six frames from each of three camera views (18 images in total). The frames are sampled at $0, 0.2, 0.4, 0.6, 0.8,$ and $1.0\text{ s}$ before termination. Instead of generating text, PROVE-Lite uses the 4B VLM backbone \cite{qwen35_2026} as a single-pass multimodal encoder. The prompt lists the task conditions $\mathcal{K}_\tau$ of the active action and all 44 terminal classes (\Cref{sec:method_rl}), and ends with a task-completion query. This schema is fixed for each action, not generated per episode as in PROVE. In one forward pass with no generated tokens, PROVE-Lite takes the final-layer hidden state $\mathbf{h} \in \mathbb{R}^{d}$ at the query position and predicts
\begin{equation}
    \big(S, \{(s_k, r_k)\}_{k \in \mathcal{K}_\tau}\big) = f_{\text{Lite}}(\mathcal{I}, I_{\text{final}}),
    \label{eq:jev_formulation}
\end{equation}
where $S \in \{0,1\}$ is the binary outcome, $s_k \in \{\texttt{SATISFIED}, \texttt{UNMET}\}$ is the binary status of condition $k$, and $r_k$ is its terminal state and failure reason class.

\paragraph{Readout Heads}
Two heads read out directly from $\mathbf{h}$:
\begin{itemize}
    \item \textbf{DecisionHead ($\mathbf{h} \to S$).}
    Inspired by verbalizer-based classification~\cite{schick2021pet}, we initialize a linear head from the pretrained language model's \texttt{false} and \texttt{true} token embeddings to predict the binary outcome $S$.

    \item \textbf{ConditionHead ($\mathbf{h} \to \{s_k, r_k, \mathbf{e}_k\}$).}
    A shared two-layer MLP predicts, for each active condition $k$, its status $s_k$, its class $r_k$ restricted to the classes valid for $k$ (44 classes across all conditions), and an auxiliary evidence vector $\mathbf{e}_k \in \{0,1\}^{18}$ indicating which of the six frames and three views support the judgment.
\end{itemize}
The evidence targets ground each condition judgment in specific frames and views during training; the reward uses only $S$, $s_k$, and $r_k$.

\paragraph{PROVE-Guided Pseudo-Labeling}
We run PROVE on training rollouts to obtain supervision without manual annotation. Related output-based distillation uses teacher-generated sequences~\cite{kim2016sequence} or rationales~\cite{hsieh2023distilling}; here, teacher outputs supervise structured classification heads instead of soft teacher logits. The binary target is PROVE's decision, $\tilde{S} = \hat{y}$. Condition targets need one more step. PROVE writes its conditions $C_3$ and its failure reason $\rho$ in free-form, task-specific language, while the heads use the fixed schema of \Cref{sec:method_rl}. A mapping VLM, \texttt{Qwen3-VL-8B-Instruct}~\cite{qwen3vl2025}, bridges the two. It receives the terminal frames $I_{\text{final}}$ alongside $C_3$, $\hat{y}$, $\rho$, and the schema for action $\tau$. For each condition $k \in \mathcal{K}_\tau$, it returns a binary status $\tilde{s}_k \in \{\texttt{SATISFIED}, \texttt{UNMET}\}$, a class $\tilde{r}_k$, and the supporting frames and views $\tilde{\mathbf{e}}_k$ directly within its structured output. Any condition not positively confirmed as satisfied is labeled \texttt{UNMET}, so silence or ambiguity from the teacher is never read as success.

\paragraph{Weighted Multi-Head Fine-Tuning}
We jointly fine-tune LoRA adapters \cite{hu2022lora} on the language backbone and both heads, keeping the vision encoder frozen, under the objective
\begin{equation}
    \mathcal{L} = w_s\,\mathcal{L}_{\text{CE}}(S, \tilde{S}) + \frac{w_r}{|\mathcal{K}_\tau|} \sum_{k \in \mathcal{K}_\tau} \mathcal{L}_k,
    \label{eq:lite_loss}
\end{equation}
\begin{equation}
    \mathcal{L}_k = \mathcal{L}_{\text{CE}}(s_k, \tilde{s}_k) + \alpha_{\tilde{r}_k}\mathcal{L}_{\text{CE}}(r_k, \tilde{r}_k) + \beta\,\mathcal{L}_{\text{BCE}}(\mathbf{e}_k, \tilde{\mathbf{e}}_k),
    \label{eq:lite_cond_loss}
\end{equation}
where $w_s$ and $w_r$ balance the binary and diagnostic terms, $\alpha_{\tilde{r}_k}$ is a class-balancing weight that counteracts the dominance of frequent classes, and $\beta = 0.3$ down-weights the auxiliary evidence term.

\subsection{Diagnosis-Guided Terminal Reward Design}
\label{sec:method_rl}

Binary terminal rewards make no distinction between failed rollouts, regardless of their physical progress. We therefore convert \textbf{PROVE-Lite}'s terminal-state diagnosis into a graded reward that grants failed rollouts partial credit for the progress they visibly achieve.

\paragraph{Task Conditions}
Each atomic action $\tau$ is assigned a small set of task conditions $\mathcal{K}_\tau$ with fixed weights $w_k$ that sum to one, \eg, \texttt{door\_closed} for \texttt{CloseFridge}, or \texttt{object\_inside} and \texttt{stable\_placement} for \texttt{PickPlaceCounterToCabinet}. Each condition admits a closed set of terminal classes, \eg, \texttt{closed}, \texttt{gap}, or \texttt{open} for \texttt{door\_closed}, from which PROVE-Lite predicts $r_k$.

\paragraph{Diagnosis-to-Severity Mapping}
We map each predicted class $r_k$ to a severity penalty $d(r_k) \in [0,1]$ that increases with its physical distance from the target configuration (\Cref{tab:rl_lookup}): a closed door incurs no penalty ($0.0$), a narrow gap a mild one ($0.3$), and an open door the maximum ($1.0$). The penalty depends only on the class and is shared by all actions that use the condition. Conditions whose target state cannot be visually confirmed are treated as unachieved ($1.0$). Conditions, weights, and penalties are fixed before policy training and are not tuned on evaluation outcomes.

\begin{table}[t]
\centering
\small
\setlength{\tabcolsep}{3pt}
\renewcommand{\arraystretch}{1.15}
\begin{tabular*}{\linewidth}{@{\extracolsep{\fill}}lllc@{}}
\toprule
\textbf{Severity tier} & \textbf{Example classes ($r$)} & \textbf{\#} & \textbf{$d(r)$} \\
\midrule
Target achieved & \texttt{closed}, \texttt{inside} & 12 & $0.0$ \\
Mild defect & \texttt{gap}, lid \texttt{offset} & 2 & $0.3$ \\
Partial progress & \texttt{boundary}, \texttt{partial} & 4 & $0.6$ \\
Unsecured object & \texttt{held} & 2 & $0.8$ \\
Unachieved/unverified & \texttt{open}, \texttt{unknown} & 24 & $1.0$ \\
\bottomrule
\end{tabular*}
\vspace{-4pt}
\caption{\textbf{Severity tiers over the 44 terminal classes.} Each class $r$ is assigned a penalty $d(r)$ on an ordinal scale of distance from the target configuration; \# gives the number of classes per tier. The full table of conditions, classes, and per-action weights is in the supplementary material.}
\label{tab:rl_lookup}
\vspace{-4pt}
\end{table}

\begin{table*}[t]
\centering
\small
\setlength{\tabcolsep}{0pt}
\renewcommand{\arraystretch}{1.08}
\begin{tabular*}{\textwidth}{@{\extracolsep{\fill}}clccc ccc@{}}
\toprule
 & & \multicolumn{2}{c}{\textbf{Specification}} & \textbf{Verification} & & & \\
\cmidrule(lr){3-4}\cmidrule(lr){5-5}
\textbf{\#} & \textbf{Evaluator} & $I_{\text{init}}, I_{\text{obj}}$ & $\hat{I}_{\text{goal}}$ & $\hat{I}_{\text{goal}}$ & \textbf{Wrong acc.\,$\downarrow$} & \textbf{Succ.\ rej.\,$\downarrow$} & \textbf{Agree.\,$\uparrow$} \\
\midrule
1 & Simulator predicate (privileged state) & & & & 22.32 & 37.16 & 69.09 \\
2 & GPT-6.1 Sol, direct judge & & & & 14.50 & 13.42 & 86.12 \\
\midrule
3 & M1 & & & & 17.02 & 17.81 & 82.52 \\
4 & M2 & $\checkmark$ & & & 18.01 & 7.81 & 87.89 \\
5 & M3 (PROVE) & $\checkmark$ & $\checkmark$ & & \underline{5.84} & \underline{1.22} & \underline{96.83} \\
\midrule
6 & M2 + goal at verification & $\checkmark$ & & $\checkmark$ & 13.95 & 18.77 & 83.26 \\
7 & M3 + goal at verification & $\checkmark$ & $\checkmark$ & $\checkmark$ & 13.26 & 6.91 & 90.42 \\
\midrule
8 & PROVE, GPT-6.1 Sol backbone & $\checkmark$ & $\checkmark$ & & \textbf{5.81} & \textbf{1.17} & \textbf{96.88} \\
9 & PROVE-Lite (Qwen3.5-4B, 1 pass) & & & & 6.21 & 10.69 & 91.20 \\
\bottomrule
\end{tabular*}
\vspace{-4pt}
\caption{\textbf{Human alignment on the 6,500-episode benchmark (65 atomic actions).}
Rates are in \%. All evaluators judge the instruction and terminal frames; checkmarks mark extra inputs.
M variants use Qwen3-VL-8B unless noted.
Specification raises agreement from 82.52 to 96.83 (rows 3--5).
The goal image helps for specification and hurts at verification (rows 6--7).
The same framework lifts GPT-6.1 Sol from 86.12 to 96.88 (rows 2, 8).
Bold: best; underline: second best.}
\label{tab:m1_m3_jev}
\vspace{-6pt}
\end{table*}

\paragraph{Graded Terminal Reward}
We aggregate the weighted penalties into a physical completion score
\begin{equation}
    C = 1 - \sum_{k \in \mathcal{K}_\tau} w_k\, d(r_k) \in [0, 1],
    \label{eq:completion}
\end{equation}
and combine it with PROVE-Lite's binary outcome $S \in \{0,1\}$:
\begin{equation}
    R = \lambda S + (1-\lambda)\, C,
    \label{eq:rl_reward}
\end{equation}
where $\lambda$ caps the partial credit a failure can earn at $1-\lambda$ of a success. Any $\lambda > 0.5$ ranks every predicted success above every predicted failure; we use $\lambda = 0.8$, so binary success remains the dominant signal. For a predicted failure ($S=0$), $C$ grades physical progress and adds at most 0.2. For a predicted success ($S=1$), $C<1$ only when the two heads disagree, \ie, the ConditionHead marks some condition as \texttt{UNMET}. Such a rollout then receives less than 1. The narrow-gap failure above ($S=0$, $d=0.3$), for instance, receives $R = 0.14$ rather than $0$. Setting $\lambda = 1$ recovers the purely binary reward $R = S$. This task-specific terminal objective is not potential-based shaping~\cite{ng1999shaping} and carries no policy-invariance guarantee. The reward is issued only at episode termination, and GAE \cite{schulman2016gae} propagates it across action chunks \cite{zhao2023act} during PPO \cite{schulman2017ppo} post-training.


\section{Experiments}
\label{sec:experiments}

We first measure human alignment and ablate each specification stage (\cref{sec:exp_main}).
We then re-evaluate open-source policies with PROVE (\cref{sec:exp_reeval}).
Finally, we test whether PROVE-Lite keeps this alignment at low cost (\cref{sec:analysis}) and whether its diagnosis improves post-training (\cref{sec:exp_rl}).


\subsection{Experimental Setup}
\label{sec:exp_setup}

\paragraph{Data}
We use three sets of rollouts.
The benchmark has 6,500 rollouts of the RoboCasa365 Xiaomi-Robotics-1 policy~\cite{xiaomi2026xr1} over 65 atomic actions, with 10 scenes per action and 10 rollouts per scene.
Humans label 3,762 successes and 2,738 failures from the instruction, the initial scene, and the terminal multi-view frames.
A second annotator independently relabels 1,000 random rollouts and matches the primary labels on 95.4\% of them.
All alignment and diagnosis results use this set.
PROVE-Lite trains on 32,500 rollouts covering all 65 actions, with 50 scenes per action and 10 rollouts per scene, using PROVE pseudo-labels only.
None of these scenes overlap with the 6,500 benchmark rollouts.
Post-training uses 15 actions, each with 80 training and 20 held-out test initializations shared by all groups.
The supplement gives the completion rubric and split details.

\paragraph{Models}
PROVE runs all stages on one frozen \texttt{Qwen3-VL-8B-Instruct} \cite{qwen3vl2025} in BF16.
Goal images come from \texttt{gpt-image-2}~\cite{openai2026gptimage2}.
PROVE-Lite uses \texttt{Qwen3.5-4B} \cite{qwen35_2026} with LoRA \cite{hu2022lora} ($r=16$) on the language model and thresholds success at 0.5.

\paragraph{Baselines}
For human alignment, we compare with native predicates and with GPT-6.1 Sol~\cite{openai2026gpt61sol} judging from the instruction and terminal frames only.
As an ablation, we also give $\hat{I}_{\text{goal}}$ to the verifier of M2 and M3.
For distillation, we LoRA-tune \texttt{Qwen3-VL-8B} and \texttt{Qwen3.5-4B} on the same pseudo-labels to generate the verdict as text.

\paragraph{Policy post-training}
All groups start from \texttt{RLinf-pi0-robocasa-pretrain-human300} \cite{zang2025rlinfvla,black2024pi0}.
The three trained groups differ only in the terminal reward.
Each trains a rank-16 policy LoRA with five PPO iterations over eight parallel environments.
PROVE-Lite stays frozen when it provides the reward.
Other hyperparameters are in the supplementary material.

\begin{figure*}[t]
\centering
\includegraphics[width=\textwidth]{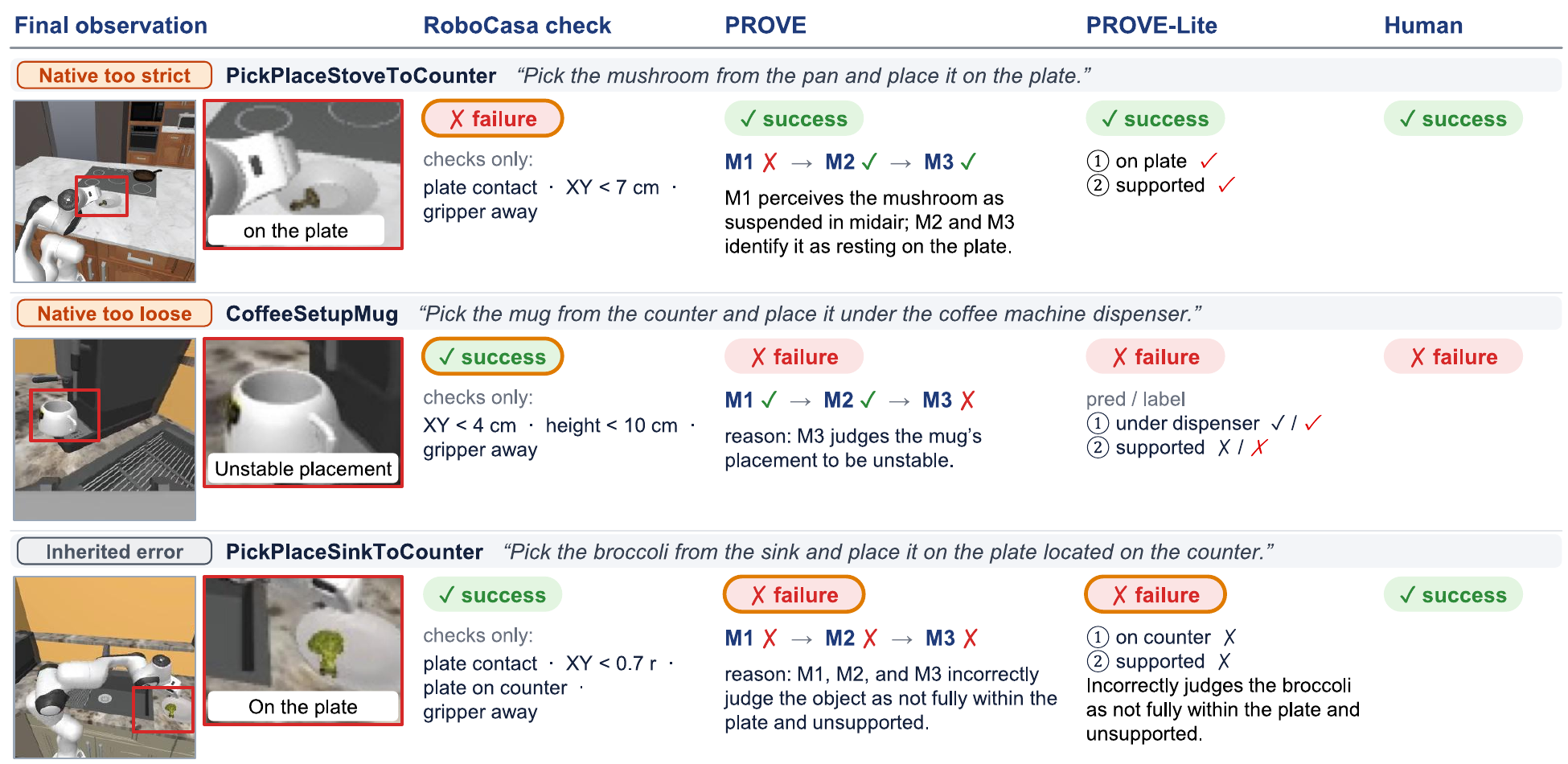}
\caption{\textbf{Qualitative comparison.}
Each row shows the final observation with a zoomed view of the evidence, followed by the verdicts of RoboCasa, PROVE, PROVE-Lite, and humans.
Orange outlines mark verdicts that disagree with human review. ``Supported'' means stably placed on and supported by the target surface.}
\label{fig:qualitative}
\vspace{-6pt}
\end{figure*}

\paragraph{Metrics}
Against human labels, we report agreement, wrong acceptance (failures judged successful), and success rejection (successes judged failed).
Diagnosis is scored against PROVE-derived condition labels, with accuracy and macro-F1 for status and class and mean absolute error for $C$ (\cref{eq:completion}).
Post-training reports PROVE-judged success as the mean and standard deviation over four seeds.

\subsection{PROVE: Human Alignment and Ablations}
\label{sec:exp_main}

\paragraph{Native predicates and direct judging fall short}
Neither native predicates nor a frontier judge matches human outcomes (\cref{tab:m1_m3_jev}, rows 1--2).
Native predicates agree on only 69.09\% of rollouts.
They accept 22.32\% of failures and reject 37.16\% of successes.
Their checks cover only part of the task, so they err in both directions.
A direct GPT-6.1 Sol judge does better at 86.12\%, but still errs on about one rollout in seven.

\paragraph{Each specification stage helps}
Agreement rises at every stage: 82.52\% for M1, 87.89\% for M2, and 96.83\% for M3 (rows 3--5).
The stages fix different errors.
Instruction-only M1 rejects 17.81\% of successes, mostly by imposing conditions the scene does not require.
Scene grounding in M2 removes most of these, cutting success rejection to 7.81\%.
Its wrong acceptance stays high at 18.01\%, because the initial scene cannot specify the final geometry.
The goal image in M3 supplies this geometry, cutting wrong acceptance to 5.84\% and success rejection to 1.22\%.
M3 is 27.74 points above native predicates and at the level of human inter-annotator agreement (95.4\%).

\paragraph{Remaining errors}
PROVE's remaining 206 errors concentrate on appliance activation, \eg, starting a kettle or a microwave, and fine closure, \eg, a refrigerator door.
These cases involve transient states, occluded contact, or millimeter-scale gaps that terminal frames show poorly.
\Cref{fig:qualitative} shows typical cases. RoboCasa rejects a valid placement and accepts a tilted mug, while PROVE rejects a valid placement that PROVE-Lite inherits.

\paragraph{The goal image specifies; it should not verify}
Using the goal image as evidence hurts both M2 and M3 (rows 6--7).
Showing $\hat{I}_{\text{goal}}$ again at verification lowers M3 from 96.83\% to 90.42\%.
Wrong acceptance rises to 13.26\%, and success rejection to 6.91\%.
Adding the goal only at verification also lowers M2, from 87.89\% to 83.26\%.
A generated goal shows one plausible outcome, not every valid one.
Used as evidence, it likely penalizes valid alternatives and distracts from the terminal frames.

\paragraph{The gain comes from the framework}
The improvement transfers across backbones (rows 2 and 8).
With GPT-6.1 Sol as its backbone, PROVE reaches 96.88\%, on par with the 8B backbone.
The same model judging directly reaches only 86.12\%.
Progressive specification thus adds 10.76 points even to a frontier model.


\begin{table}[t]
\centering
\footnotesize
\setlength{\tabcolsep}{0pt}
\renewcommand{\arraystretch}{1.08}
\begin{tabular*}{\columnwidth}{@{\extracolsep{\fill}}lcccc@{}}
\toprule
& \multicolumn{2}{c}{Success (\%) / rank}
& & Human support \\
\cmidrule(lr){2-3}
Model & Native & PROVE & Disagree & N / P \\
\midrule
GR00T N1.5
& 53.06\,{\scriptsize(4)} & 41.12\,{\scriptsize(3)}
& 842 & 44 / 798 \\
RLDX-1
& 69.98\,{\scriptsize(2)} & 40.11\,{\scriptsize(4)}
& 2,242 & 117 / 2,125 \\
PRTS
& 66.08\,{\scriptsize(3)} & 43.91\,{\scriptsize(2)}
& 2,637 & 109 / 2,528 \\
Awomo-0.5I
& 86.75\,{\scriptsize(1)} & 46.29\,{\scriptsize(1)}
& 2,956 & 127 / 2,829 \\
\midrule
All & -- & -- & 8,677 & 397 / 8,280 \\
\bottomrule
\end{tabular*}
\vspace{-4pt}
\caption{\textbf{Policy re-evaluation.}
Each policy runs on the 6,500 benchmark initializations (\protect\cref{sec:exp_setup}); ranks are in parentheses.
Disagree counts rollouts where Native and PROVE differ.
On these, annotators blind to both labels side with Native (N) or PROVE (P); 95.42\% side with PROVE.}
\label{tab:policy_reevaluation}
\vspace{-6pt}
\end{table}

\subsection{Re-evaluating Open-Source Policies}
\label{sec:exp_reeval}

Native success overstates progress and misorders policies (\cref{tab:policy_reevaluation}). We test three vision-language-action (VLA) policies, GR00T N1.5~\cite{bjorck2025gr00tn1,nvidia2025gr00tn15}, RLDX-1~\cite{kim2026rldx1}, and PRTS~\cite{zhang2026prts}, and one world action model (WAM), Awomo-0.5I~\cite{awomo2026}. Native predicates report 12 to 40 points higher success than PROVE. To decide which judgment is right, annotators blind to both labels review all 8,677 disagreements. They side with PROVE in 95.42\% of them. The ranking changes as well. RLDX-1 falls from second to last, and PRTS rises from third to second. Awomo-0.5I stays first, but its lead shrinks from 16.77 to 2.38 points.

\begin{table}[t]
\centering
\small
\setlength{\tabcolsep}{0pt}
\renewcommand{\arraystretch}{1.12}
\begin{tabular*}{\linewidth}{@{\extracolsep{\fill}}lcccc@{}}
\toprule
\textbf{Evaluator} & \textbf{Gen.\ tokens} & \textbf{Agree.} & \textbf{Lat.\,(s)} & \textbf{Mem.\,(GB)} \\
\midrule
PROVE (M3) & 386 & 96.83 & 10.21 & 16.95 \\
\midrule
Generative 8B & 59.2 & 91.40 & 1.03 & 16.78 \\
Generative 4B & 33.8 & 89.20 & 0.84 & 9.04 \\
PROVE-Lite, untrained & 0 & 56.80 & 0.43 & 8.96 \\
\textbf{PROVE-Lite} & 0 & 91.20 & \textbf{0.27} & 8.97 \\
\bottomrule
\end{tabular*}
\vspace{-4pt}
\caption{\textbf{Distillation and efficiency.}
Agreement in \%; latency per episode and peak memory on one NVIDIA RTX PRO 4500 GPU.
Generative evaluators are LoRA-tuned on PROVE's pseudo-labels; the untrained PROVE-Lite uses the verbalizer-initialized head.}
\label{tab:efficiency}
\vspace{-6pt}
\end{table}

\begin{table}[t]
\centering
\small
\setlength{\tabcolsep}{0pt}
\renewcommand{\arraystretch}{1.15}
\begin{tabular*}{\linewidth}{@{\extracolsep{\fill}}llccc@{}}
\toprule
\textbf{Prediction} & \textbf{Space} & \textbf{Acc.\,(\%)} & \textbf{Macro-F1} & \textbf{MAE} \\
\midrule
Status $s_k$ & 2 per cond. & 87.80 & 0.817 & -- \\
Class $r_k$ & 44 total & 80.77 & 0.542 & -- \\
Completion $C$ & $[0,1]$ & -- & -- & 0.165 \\
\bottomrule
\end{tabular*}
\vspace{-4pt}
\caption{\textbf{PROVE-Lite diagnosis against PROVE-derived condition labels}, over active conditions on the benchmark.}
\label{tab:stage2_detail}
\vspace{-8pt}
\end{table}

\subsection{PROVE-Lite: Distillation and Diagnosis}
\label{sec:analysis}

\paragraph{Distillation keeps most of the alignment}
Without reference images, PROVE-Lite agrees with humans on 91.20\% of rollouts across all 65 actions (\cref{tab:m1_m3_jev}, row 9).
Its wrong acceptance is 6.21\%, close to the 5.84\% of PROVE, while its success rejection is higher at 10.69\%.
Since no benchmark scene appears in training, this measures generalization to new layouts and styles.
Unlike PROVE, it never sees the initial scene.
Actions judged against the initial state, such as lowering the heat in \cref{fig:native_false_success}b, therefore exceed its input.

\paragraph{A single readout is much cheaper than generation}
PROVE-Lite matches the generative 8B evaluator at about a quarter of its latency (\cref{tab:efficiency}).
PROVE-Lite reaches 91.20\%, against 91.40\% for the generative 8B and 89.20\% for the generative 4B.
It needs 0.27\,s per episode, against 1.03\,s and 0.84\,s.
It is about 38$\times$ faster than PROVE and uses about half its memory.
Without distillation, the same model reaches only 56.80\%, so the alignment comes from training, not from the readout alone.
Removing decoding makes PROVE-Lite cheap enough for the training loop.

\paragraph{The diagnosis is accurate enough to grade failures}
Against PROVE-derived condition labels (\cref{tab:stage2_detail}), status accuracy is 87.80\% (0.817 macro-F1).
Terminal-class accuracy is 80.77\%, with 0.542 macro-F1 over 44 classes.
The predicted completion score $C$ has a mean absolute error of 0.165.
With weight $1-\lambda=0.2$ in \cref{eq:rl_reward}, this amounts to only about 0.033 in reward.
For a predicted success, the condition term lowers $R$ only when the two heads disagree.
This happens for 10.8\% of predicted successes, whose mean reward is then 0.877.
A false binary acceptance still scores at least $\lambda=0.8$, above any predicted failure.

\begin{table}[t]
\centering
\small
\setlength{\tabcolsep}{0pt}
\renewcommand{\arraystretch}{1.15}
\begin{tabular*}{\linewidth}{@{\extracolsep{\fill}}llc@{}}
\toprule
\textbf{Policy group} & \textbf{Terminal reward} & \textbf{PROVE success (\%)} \\
\midrule
Base policy & -- & $50.17 \pm 0.77$ \\
Simulator & simulator binary & $52.00 \pm 0.81$ \\
Binary & $R = S$ & $54.58 \pm 0.88$ \\
\textbf{Graded (ours)} & $R = \lambda S + (1-\lambda)C$ & $\mathbf{56.58 \pm 0.91}$ \\
\bottomrule
\end{tabular*}
\vspace{-4pt}
\caption{\textbf{PPO post-training on 15 tasks.}
Mean $\pm$ s.d.\ over four seeds, with 20 test episodes per task and seed.
All groups share the initialization and training budget; Binary and Graded use PROVE-Lite rewards with $\lambda = 0.8$ (\protect\cref{eq:rl_reward}).}
\label{tab:rl_groups}
\vspace{-8pt}
\end{table}

\subsection{Diagnosis-Guided Policy Post-Training}
\label{sec:exp_rl}

A more reliable reward helps, and grading failures by severity helps further (\cref{tab:rl_groups}).
All groups start from one checkpoint and share the same 1,200 held-out test episodes.
The binary PROVE-Lite reward beats the simulator reward by 2.58 points.
The graded reward adds 2.00 more, reaching 56.58\%, 6.41 points above the base policy.
Its gain over the binary reward holds across seeds (paired $t$-test, $p=0.031$).
To avoid scoring policies with their own reward model, PROVE judges the test rollouts.
It differs from PROVE-Lite in inputs, model size, and access to reference assets.

\section{Conclusion}
\label{sec:conclusion}

We presented \textbf{PROVE}, an evaluator for open-world robot manipulation built on one principle.
Predictions specify what success should look like, and only real observations verify whether it happened.
Each grounding stage, from the task to the scene to a generated goal image, sharpens the success conditions.
Across 65 RoboCasa365 actions, this raises human agreement from 82.52\% to \textbf{96.83\%}.
Re-evaluated with PROVE, open-source policies score far below their native success, and their ranking changes.
Distilled into \textbf{PROVE-Lite}, the same judgment becomes cheap enough for the training loop.
PROVE-Lite keeps \textbf{91.20\%} agreement in a single forward pass and diagnoses the terminal state of each condition.
Mapped to fixed severities, these diagnoses yield a graded reward that beats simulator and binary rewards in PPO.
This suggests that knowing \emph{how} an execution failed, not only \emph{whether}, is a useful learning signal.

\paragraph{Limitations and future work}
Our experiments use one simulator, RoboCasa365. PROVE needs no simulator-specific training, so applying it to other benchmarks and to physical robots is a natural next step.
Task conditions and severities are defined per action; learning them from data would extend the reward to new tasks.
Scaling diagnosis-guided post-training to more tasks and longer budgets is another promising direction.

\clearpage
{
\small
\bibliographystyle{ieeenat_fullname}
\bibliography{references}

@inproceedings{robocasa2024,
  title        = {{RoboCasa}: Large-Scale Simulation of Household Tasks for Generalist Robots},
  author       = {Nasiriany, Soroush and Maddukuri, Abhiram and Zhang, Lance and Parikh, Adeet and Lo, Aaron and Joshi, Abhishek and Mandlekar, Ajay and Zhu, Yuke},
  booktitle    = {Robotics: Science and Systems (RSS)},
  year         = {2024},
  url          = {https://roboticsproceedings.org/rss20/p050.html},
  doi          = {10.15607/RSS.2024.XX.050}
}

@inproceedings{robocasa365_2026,
  title        = {{RoboCasa365}: A Large-Scale Simulation Framework for Training and Benchmarking Generalist Robots},
  author       = {Nasiriany, Soroush and Nasiriany, Sepehr and Maddukuri, Abhiram and Zhu, Yuke},
  booktitle    = {International Conference on Learning Representations (ICLR)},
  year         = {2026},
  url          = {https://openreview.net/forum?id=tQJYKwc3n4}
}

@article{qwen3vl2025,
  title        = {{Qwen3-VL} Technical Report},
  author       = {Bai, Shuai and Cai, Yuxuan and Chen, Ruizhe and Chen, Keqin and Chen, Xionghui and Cheng, Zesen and Deng, Lianghao and Ding, Wei and Gao, Chang and Ge, Chunjiang and others},
  journal      = {arXiv preprint arXiv:2511.21631},
  year         = {2025},
  url          = {https://arxiv.org/abs/2511.21631}
}

@misc{qwen35_2026,
  title        = {{Qwen3.5}: Towards Native Multimodal Agents},
  author       = {{Qwen Team}},
  month        = {feb},
  year         = {2026},
  howpublished = {\url{https://qwen.ai/blog?id=qwen3.5}}
}

@inproceedings{schick2021pet,
  title        = {Exploiting Cloze-Questions for Few-Shot Text Classification and Natural Language Inference},
  author       = {Schick, Timo and Sch{\"u}tze, Hinrich},
  booktitle    = {Proceedings of the 16th Conference of the European Chapter of the Association for Computational Linguistics (EACL)},
  pages        = {255--269},
  year         = {2021},
  url          = {https://aclanthology.org/2021.eacl-main.20/},
  doi          = {10.18653/v1/2021.eacl-main.20}
}

@inproceedings{zhou2025autoeval,
  title        = {{AutoEval}: Autonomous Evaluation of Generalist Robot Manipulation Policies in the Real World},
  author       = {Zhou, Zhiyuan and Atreya, Pranav and Tan, You Liang and Pertsch, Karl and Levine, Sergey},
  booktitle    = {Conference on Robot Learning (CoRL)},
  series       = {Proceedings of Machine Learning Research},
  volume       = {305},
  pages        = {1997--2017},
  year         = {2025},
  url          = {https://proceedings.mlr.press/v305/zhou25a.html}
}

@inproceedings{zheng2023judging,
  title        = {Judging {LLM}-as-a-Judge with {MT-Bench} and {Chatbot Arena}},
  author       = {Zheng, Lianmin and Chiang, Wei-Lin and Sheng, Ying and Zhuang, Siyuan and Wu, Zhanghao and Zhuang, Yonghao and Lin, Zi and Li, Zhuohan and Li, Dacheng and Xing, Eric P. and Zhang, Hao and Gonzalez, Joseph E. and Stoica, Ion},
  booktitle    = {Advances in Neural Information Processing Systems (NeurIPS), Datasets and Benchmarks Track},
  year         = {2023},
  url          = {https://arxiv.org/abs/2306.05685}
}

@inproceedings{du2023successvqa,
  title        = {Vision-Language Models as Success Detectors},
  author       = {Du, Yuqing and Konyushkova, Ksenia and Denil, Misha and Raju, Akhil and Landon, Jessica and Hill, Felix and de Freitas, Nando and Cabi, Serkan},
  booktitle    = {Proceedings of The 2nd Conference on Lifelong Learning Agents (CoLLAs)},
  series       = {Proceedings of Machine Learning Research},
  volume       = {232},
  year         = {2023},
  url          = {https://proceedings.mlr.press/v232/du23b.html},
  pages        = {120--136}
}

@inproceedings{sermanet2024robovqa,
  title        = {{RoboVQA}: Multimodal Long-Horizon Reasoning for Robotics},
  author       = {Sermanet, Pierre and Ding, Tianli and Zhao, Jeffrey and Xia, Fei and Dwibedi, Debidatta and Gopalakrishnan, Keerthana and Chan, Christine and Dulac-Arnold, Gabriel and Maddineni, Sharath and Joshi, Nikhil J. and others},
  booktitle    = {IEEE International Conference on Robotics and Automation (ICRA)},
  year         = {2024},
  url          = {https://arxiv.org/abs/2311.00899}
}

@inproceedings{black2024susie,
  title        = {Zero-Shot Robotic Manipulation with Pretrained Image-Editing Diffusion Models},
  author       = {Black, Kevin and Nakamoto, Mitsuhiko and Atreya, Pranav and Walke, Homer and Finn, Chelsea and Kumar, Aviral and Levine, Sergey},
  booktitle    = {International Conference on Learning Representations (ICLR)},
  year         = {2024},
  url          = {https://arxiv.org/abs/2310.10639}
}

@inproceedings{du2023unipi,
  title        = {Learning Universal Policies via Text-Guided Video Generation},
  author       = {Du, Yilun and Yang, Mengjiao and Dai, Bo and Dai, Hanjun and Nachum, Ofir and Tenenbaum, Joshua B. and Schuurmans, Dale and Abbeel, Pieter},
  booktitle    = {Advances in Neural Information Processing Systems (NeurIPS)},
  year         = {2023},
  url          = {https://arxiv.org/abs/2302.00111}
}

@inproceedings{rocamonde2024vlmrm,
  title        = {Vision-Language Models are Zero-Shot Reward Models for Reinforcement Learning},
  author       = {Rocamonde, Juan and Montesinos, Victoriano and Nava, Elvis and Perez, Ethan and Lindner, David},
  booktitle    = {International Conference on Learning Representations (ICLR)},
  year         = {2024},
  url          = {https://arxiv.org/abs/2310.12921}
}

@article{baumli2023vlmrewards,
  title        = {Vision-Language Models as a Source of Rewards},
  author       = {Baumli, Kate and Baveja, Satinder and Behbahani, Feryal and Chan, Harris and Comanici, Gheorghe and Flennerhag, Sebastian and Gazeau, Maxime and Holsheimer, Kristian and Horgan, Dan and Laskin, Michael and others},
  journal      = {arXiv preprint arXiv:2312.09187},
  year         = {2023},
  url          = {https://arxiv.org/abs/2312.09187}
}

@inproceedings{wang2024rlvlmf,
  title        = {{RL-VLM-F}: Reinforcement Learning from Vision Language Foundation Model Feedback},
  author       = {Wang, Yufei and Sun, Zhanyi and Zhang, Jesse and Xian, Zhou and B{\i}y{\i}k, Erdem and Held, David and Erickson, Zackory},
  booktitle    = {Proceedings of the 41st International Conference on Machine Learning (ICML)},
  series       = {Proceedings of Machine Learning Research},
  volume       = {235},
  year         = {2024},
  url          = {https://proceedings.mlr.press/v235/wang24bn.html},
  pages        = {51484--51501}
}

@inproceedings{ma2024eureka,
  title        = {{Eureka}: Human-Level Reward Design via Coding Large Language Models},
  author       = {Ma, Yecheng Jason and Liang, William and Wang, Guanzhi and Huang, De-An and Bastani, Osbert and Jayaraman, Dinesh and Zhu, Yuke and Fan, Linxi and Anandkumar, Anima},
  booktitle    = {International Conference on Learning Representations (ICLR)},
  year         = {2024},
  url          = {https://arxiv.org/abs/2310.12931}
}

@inproceedings{ma2025gvl,
  title        = {Vision Language Models are In-Context Value Learners},
  author       = {Ma, Yecheng Jason and Hejna, Joey and Wahid, Ayzaan and Fu, Chuyuan and Shah, Dhruv and Liang, Jacky and Xu, Zhuo and Kirmani, Sean and Xu, Peng and Driess, Danny and others},
  booktitle    = {International Conference on Learning Representations (ICLR)},
  year         = {2025},
  url          = {https://arxiv.org/abs/2411.04549}
}

@inproceedings{liang2026robometer,
  title        = {{Robometer}: Scaling General-Purpose Robotic Reward Models via Trajectory Comparisons},
  author       = {Liang, Anthony and Korkmaz, Yigit and Zhang, Jiahui and Hwang, Minyoung and Anwar, Abrar and Kaushik, Sidhant and Shah, Aditya and Huang, Alex S. and Zettlemoyer, Luke and Fox, Dieter and Xiang, Yu and Li, Anqi and Bobu, Andreea and Gupta, Abhishek and Tu, Stephen and B{\i}y{\i}k, Erdem and Zhang, Jesse},
  booktitle    = {Robotics: Science and Systems (RSS)},
  year         = {2026},
  url          = {https://arxiv.org/abs/2603.02115}
}

@article{procvlm2026,
  title        = {{ProcVLM}: Learning Procedure-Grounded Progress Rewards for Robotic Manipulation},
  author       = {Feng, Youhe and Shi, Hansen and Li, Haoyang and Guo, Xinlei and Wang, Yang and Zhang, Chengyang and Zhang, Jinkai and Zhang, Xiaohan and Tang, Jie and Zhang, Jing},
  journal      = {arXiv preprint arXiv:2605.08774},
  year         = {2026},
  url          = {https://arxiv.org/abs/2605.08774}
}

@inproceedings{brooks2023instructpix2pix,
  title        = {{InstructPix2Pix}: Learning to Follow Image Editing Instructions},
  author       = {Brooks, Tim and Holynski, Aleksander and Efros, Alexei A.},
  booktitle    = {Proceedings of the IEEE/CVF Conference on Computer Vision and Pattern Recognition (CVPR)},
  year         = {2023},
  url          = {https://arxiv.org/abs/2211.09800}
}

@article{hinton2015distilling,
  title        = {Distilling the Knowledge in a Neural Network},
  author       = {Hinton, Geoffrey and Vinyals, Oriol and Dean, Jeff},
  journal      = {arXiv preprint arXiv:1503.02531},
  year         = {2015},
  url          = {https://arxiv.org/abs/1503.02531}
}

@inproceedings{hu2022lora,
  title        = {{LoRA}: Low-Rank Adaptation of Large Language Models},
  author       = {Hu, Edward J. and Shen, Yelong and Wallis, Phillip and Allen-Zhu, Zeyuan and Li, Yuanzhi and Wang, Shean and Wang, Lu and Chen, Weizhu},
  booktitle    = {International Conference on Learning Representations (ICLR)},
  year         = {2022},
  url          = {https://arxiv.org/abs/2106.09685}
}

@article{schulman2017ppo,
  title        = {Proximal Policy Optimization Algorithms},
  author       = {Schulman, John and Wolski, Filip and Dhariwal, Prafulla and Radford, Alec and Klimov, Oleg},
  journal      = {arXiv preprint arXiv:1707.06347},
  year         = {2017},
  url          = {https://arxiv.org/abs/1707.06347}
}

@inproceedings{schulman2016gae,
  title        = {High-Dimensional Continuous Control Using Generalized Advantage Estimation},
  author       = {Schulman, John and Moritz, Philipp and Levine, Sergey and Jordan, Michael and Abbeel, Pieter},
  booktitle    = {International Conference on Learning Representations (ICLR)},
  year         = {2016},
  url          = {https://arxiv.org/abs/1506.02438}
}

@inproceedings{zhao2023act,
  title        = {Learning Fine-Grained Bimanual Manipulation with Low-Cost Hardware},
  author       = {Zhao, Tony Z. and Kumar, Vikash and Levine, Sergey and Finn, Chelsea},
  booktitle    = {Robotics: Science and Systems (RSS)},
  year         = {2023},
  url          = {https://arxiv.org/abs/2304.13705}
}

@article{black2024pi0,
  title        = {$\pi_0$: A Vision-Language-Action Flow Model for General Robot Control},
  author       = {Black, Kevin and Brown, Noah and Driess, Danny and Esmail, Adnan and Equi, Michael and Finn, Chelsea and Fusai, Niccolo and Groom, Lachy and Hausman, Karol and Ichter, Brian and others},
  journal      = {arXiv preprint arXiv:2410.24164},
  year         = {2024},
  url          = {https://arxiv.org/abs/2410.24164}
}

@inproceedings{zang2025rlinfvla,
  title        = {{RLinf-VLA}: A Unified and Efficient Framework for Reinforcement Learning of Vision-Language-Action Models},
  author       = {Zang, Hongzhi and Wei, Mingjie and Xu, Si and Wu, Yongji and Guo, Zhen and Wang, Yuanqing and Lin, Hao and Wang, Peihong and Shi, Liangzhi and Xie, Yuqing and others},
  booktitle    = {Robotics: Science and Systems (RSS)},
  year         = {2026},
  url          = {https://arxiv.org/abs/2510.06710}
}

@article{bjorck2025gr00tn1,
  title        = {{GR00T N1}: An Open Foundation Model for Generalist Humanoid Robots},
  author       = {{NVIDIA} and Bjorck, Johan and Casta{\~n}eda, Fernando and Cherniadev, Nikita and Da, Xingye and Ding, Runyu and Fan, Linxi and Fang, Yu and Fox, Dieter and Hu, Fengyuan and others},
  journal      = {arXiv preprint arXiv:2503.14734},
  year         = {2025},
  url          = {https://arxiv.org/abs/2503.14734}
}

@misc{nvidia2025gr00tn15,
  title        = {{GR00T N1.5}: An Improved Open Foundation Model for Generalist Humanoid Robots},
  author       = {{NVIDIA GEAR Team}},
  year         = {2025},
  howpublished = {\url{https://research.nvidia.com/labs/gear/gr00t-n1_5/}}
}

@article{kim2026rldx1,
  title        = {{RLDX-1} Technical Report},
  author       = {Kim, Dongyoung and Jang, Huiwon and Koo, Myungkyu and Jang, Suhyeok and Kim, Taeyoung and Kim, Beomjun and Yoon, Byungjun and Jang, Changsung and Choi, Daewon and Han, Dongsu and others},
  journal      = {arXiv preprint arXiv:2605.03269},
  year         = {2026},
  url          = {https://arxiv.org/abs/2605.03269}
}

@article{zhang2026prts,
  title        = {{PRTS}: A Primitive Reasoning and Tasking System via Contrastive Representations},
  author       = {Zhang, Yang and Zhao, Jiangyuan and Fan, Chenyou and Yan, Fangzheng and Li, Tian and Tang, Haitong and Fu, Sen and Wu, Xuan'er and Weng, Qizhen and Zhang, Weinan and Li, Xiu and Zhang, Chi and Bai, Chenjia and Li, Xuelong},
  journal      = {arXiv preprint arXiv:2604.27472},
  year         = {2026},
  url          = {https://arxiv.org/abs/2604.27472}
}

@misc{awomo2026,
  title        = {{Awomo-0.5} Technical Report},
  author       = {{Awomo-WAM Team}},
  year         = {2026},
  howpublished = {Technical report. \url{https://github.com/Awomo-WestlakeDI/Awomo-0.5/blob/main/Awomo_0_5.pdf}},
  url          = {https://github.com/Awomo-WestlakeDI/Awomo-0.5/blob/main/Awomo_0_5.pdf}
}

@misc{openai2026gpt61sol,
  title        = {Introducing {GPT-6.1 Sol}},
  author       = {{OpenAI}},
  year         = {2026},
  howpublished = {\url{https://openai.com/index/introducing-gpt-6-1-sol/}},
  note         = {Accessed: 2026-11-17}
}

@misc{openai2026gptimage2,
  title        = {Introducing {ChatGPT Images 2.0}},
  author       = {{OpenAI}},
  year         = {2026},
  howpublished = {\url{https://openai.com/index/introducing-chatgpt-images-2-0/}},
  note         = {Model: gpt-image-2. Accessed: 2026-11-17}
}

@inproceedings{liu2023reflect,
  title        = {{REFLECT}: Summarizing Robot Experiences for Failure Explanation and Correction},
  author       = {Liu, Zeyi and Bahety, Arpit and Song, Shuran},
  booktitle    = {Conference on Robot Learning (CoRL)},
  year         = {2023},
  url          = {https://proceedings.mlr.press/v229/liu23g.html},
  series       = {Proceedings of Machine Learning Research},
  volume       = {229},
  pages        = {3468--3484}
}

@inproceedings{duan2025aha,
  title        = {{AHA}: A Vision-Language-Model for Detecting and Reasoning over Failures in Robotic Manipulation},
  author       = {Duan, Jiafei and Pumacay, Wilbert and Kumar, Nishanth and Wang, Yi Ru and Tian, Shulin and Yuan, Wentao and Krishna, Ranjay and Fox, Dieter and Mandlekar, Ajay and Guo, Yijie},
  booktitle    = {International Conference on Learning Representations (ICLR)},
  year         = {2025},
  url          = {https://arxiv.org/abs/2410.00371}
}

@article{pacaud2026guardian,
  title        = {{Guardian}: Detecting Robotic Manipulation Failures with Vision-Language Models},
  author       = {Pacaud, Paul and Garcia, Ricardo and Chen, Shizhe and Schmid, Cordelia},
  journal      = {arXiv preprint arXiv:2512.01946},
  year         = {2026},
  url          = {https://arxiv.org/abs/2512.01946v6},
  note         = {Version 6}
}

@inproceedings{srivastava2022behavior,
  title        = {{BEHAVIOR}: Benchmark for Everyday Household Activities in Virtual, Interactive, and Ecological Environments},
  author       = {Srivastava, Sanjana and Li, Chengshu and Lingelbach, Michael and Mart\'in-Mart\'in, Roberto and Xia, Fei and Vainio, Kent Elliott and Lian, Zheng and Gokmen, Cem and Buch, Shyamal and Liu, Karen and Savarese, Silvio and Gweon, Hyowon and Wu, Jiajun and Fei-Fei, Li},
  booktitle    = {Proceedings of the 5th Conference on Robot Learning},
  pages        = {477--490},
  year         = {2022},
  volume       = {164},
  series       = {Proceedings of Machine Learning Research},
  url          = {https://proceedings.mlr.press/v164/srivastava22a.html}
}

@inproceedings{li2023behavior1k,
  title        = {{BEHAVIOR-1K}: A Benchmark for Embodied AI with 1,000 Everyday Activities and Realistic Simulation},
  author       = {Li, Chengshu and Zhang, Ruohan and Wong, Josiah and Gokmen, Cem and Srivastava, Sanjana and Mart\'in-Mart\'in, Roberto and Wang, Chen and Levine, Gabrael and Lingelbach, Michael and Sun, Jiankai and Anvari, Mona and Hwang, Minjune and Sharma, Manasi and Aydin, Arman and Bansal, Dhruva and Hunter, Samuel and Kim, Kyu-Young and Lou, Alan and Matthews, Caleb R and Villa-Renteria, Ivan and Tang, Jerry Huayang and Tang, Claire and Xia, Fei and Savarese, Silvio and Gweon, Hyowon and Liu, Karen and Wu, Jiajun and Fei-Fei, Li},
  booktitle    = {Proceedings of The 6th Conference on Robot Learning},
  pages        = {80--93},
  year         = {2023},
  volume       = {205},
  series       = {Proceedings of Machine Learning Research},
  url          = {https://proceedings.mlr.press/v205/li23a.html}
}

@inproceedings{ma2023liv,
  title        = {{LIV}: Language-Image Representations and Rewards for Robotic Control},
  author       = {Ma, Yecheng Jason and Kumar, Vikash and Zhang, Amy and Bastani, Osbert and Jayaraman, Dinesh},
  booktitle    = {Proceedings of the 40th International Conference on Machine Learning},
  pages        = {23301--23320},
  year         = {2023},
  volume       = {202},
  series       = {Proceedings of Machine Learning Research},
  url          = {https://proceedings.mlr.press/v202/ma23b.html}
}

@inproceedings{huang2023voxposer,
  title        = {{VoxPoser}: Composable 3D Value Maps for Robotic Manipulation with Language Models},
  author       = {Huang, Wenlong and Wang, Chen and Zhang, Ruohan and Li, Yunzhu and Wu, Jiajun and Fei-Fei, Li},
  booktitle    = {Proceedings of The 7th Conference on Robot Learning},
  pages        = {540--562},
  year         = {2023},
  volume       = {229},
  series       = {Proceedings of Machine Learning Research},
  url          = {https://proceedings.mlr.press/v229/huang23b.html}
}

@inproceedings{kim2016sequence,
  title        = {Sequence-Level Knowledge Distillation},
  author       = {Kim, Yoon and Rush, Alexander M.},
  booktitle    = {Proceedings of the 2016 Conference on Empirical Methods in Natural Language Processing},
  year         = {2016},
  url          = {https://aclanthology.org/D16-1139/},
  doi          = {10.18653/v1/D16-1139},
  pages        = {1317--1327}
}

@inproceedings{hsieh2023distilling,
  title        = {Distilling Step-by-Step! Outperforming Larger Language Models with Less Training Data and Smaller Model Sizes},
  author       = {Hsieh, Cheng-Yu and Li, Chun-Liang and Yeh, Chih-kuan and Nakhost, Hootan and Fujii, Yasuhisa and Ratner, Alex and Krishna, Ranjay and Lee, Chen-Yu and Pfister, Tomas},
  booktitle    = {Findings of the {A}ssociation for {C}omputational {L}inguistics: {ACL} 2023},
  year         = {2023},
  url          = {https://aclanthology.org/2023.findings-acl.507/},
  doi          = {10.18653/v1/2023.findings-acl.507},
  pages        = {8003--8017}
}

@article{james2020rlbench,
  title        = {{RLBench}: The Robot Learning Benchmark \& Learning Environment},
  author       = {James, Stephen and Ma, Zicong and Rovick Arrojo, David and Davison, Andrew J.},
  journal      = {IEEE Robotics and Automation Letters},
  year         = {2020},
  url          = {https://arxiv.org/abs/1909.12271}
}

@inproceedings{liu2023libero,
  title        = {{LIBERO}: Benchmarking Knowledge Transfer for Lifelong Robot Learning},
  author       = {Liu, Bo and Zhu, Yifeng and Gao, Chongkai and Feng, Yihao and Liu, Qiang and Zhu, Yuke and Stone, Peter},
  booktitle    = {Advances in Neural Information Processing Systems (NeurIPS), Datasets and Benchmarks Track},
  year         = {2023},
  url          = {https://arxiv.org/abs/2306.03310}
}

@article{lee2026roboreward,
  title        = {{RoboReward}: General-Purpose Vision-Language Reward Models for Robotics},
  author       = {Lee, Tony and Wagenmaker, Andrew and Pertsch, Karl and Liang, Percy and Levine, Sergey and Finn, Chelsea},
  journal      = {arXiv preprint arXiv:2601.00675},
  year         = {2026},
  url          = {https://arxiv.org/abs/2601.00675}
}

@inproceedings{sontakke2023roboclip,
  title        = {{RoboCLIP}: One Demonstration is Enough to Learn Robot Policies},
  author       = {Sontakke, Sumedh A and Zhang, Jesse and Arnold, S{\'e}bastien M. R. and Pertsch, Karl and B{\i}y{\i}k, Erdem and Sadigh, Dorsa and Finn, Chelsea and Itti, Laurent},
  booktitle    = {Advances in Neural Information Processing Systems (NeurIPS)},
  year         = {2023},
  url          = {https://proceedings.neurips.cc/paper_files/paper/2023/hash/ae54ce310476218f26dd48c1626d5187-Abstract-Conference.html}
}

@article{bharadhwaj2024gen2act,
  title        = {{Gen2Act}: Human Video Generation in Novel Scenarios Enables Generalizable Robot Manipulation},
  author       = {Bharadhwaj, Homanga and Dwibedi, Debidatta and Gupta, Abhinav and Tulsiani, Shubham and Doersch, Carl and Xiao, Ted and Shah, Dhruv and Xia, Fei and Sadigh, Dorsa and Kirmani, Sean},
  journal      = {arXiv preprint arXiv:2409.16283},
  year         = {2024},
  url          = {https://arxiv.org/abs/2409.16283}
}

@inproceedings{ng1999shaping,
  title        = {Policy Invariance under Reward Transformations: Theory and Application to Reward Shaping},
  author       = {Ng, Andrew Y. and Harada, Daishi and Russell, Stuart},
  booktitle    = {Proceedings of the Sixteenth International Conference on Machine Learning (ICML)},
  year         = {1999},
  pages        = {278--287},
  url          = {https://people.eecs.berkeley.edu/~pabbeel/cs287-fa09/readings/NgHaradaRussell-shaping-ICML1999.pdf}
}

@article{xiaomi2026xr1,
  title   = {{Xiaomi-Robotics-1}: Scaling Vision-Language-Action
             Models with over 100K Hours of Real-World Trajectories},
  author  = {{Xiaomi Robotics Team} and
             Guo, Jun and Jin, Piaopiao and Li, Jason and
             Li, Peiyan and Li, Yingyan and Liu, Futeng and
             Peng, Wanli and Qin, Optimus and Su, Yifei and others},
  journal = {arXiv preprint arXiv:2607.15330},
  year    = {2026},
  url     = {https://arxiv.org/abs/2607.15330}
}

@article{chen2026robojev,
  title   = {{RoboJev}: An Open Vision-Language Judge for Robot Success Detection},
  author  = {Chen, Josef Liyanjun},
  journal = {arXiv preprint arXiv:2610.33718},
  year    = {2026},
  url     = {https://arxiv.org/abs/2610.33718}
}

@misc{autotrust2026jev27bvl,
  title        = {{JEV-27B-VL}},
  author       = {{AutoTrust AI Lab}},
  year         = {2026},
  howpublished = {Hugging Face model card. \url{https://huggingface.co/autotrust/JEV-27B-VL}},
  note         = {Accessed: 2026-11-17}
}

@article{liu2026evalactions,
  title   = {{Eval-Actions}: Fine-Grained Execution Quality Evaluation for Robotic Manipulation},
  author  = {Liu, Mengyuan and Sheng, Juyi and Li, Peiming and
             Wang, Ziyi and Xu, Tianming and Xu, Tiantian and Liu, Hong},
  journal = {arXiv preprint arXiv:2601.18723},
  year    = {2026},
  url     = {https://arxiv.org/abs/2601.18723v2}
}
}

\clearpage
\appendix
\setcounter{section}{0}
\setcounter{table}{0}
\setcounter{figure}{0}
\setcounter{equation}{0}
\renewcommand{\thesection}{\Alph{section}}
\renewcommand{\thetable}{S\arabic{table}}
\renewcommand{\thefigure}{S\arabic{figure}}
\renewcommand{\theequation}{S\arabic{equation}}
\renewcommand{\theHsection}{A\arabic{section}}
\renewcommand{\theHtable}{S\arabic{table}}
\renewcommand{\theHfigure}{S\arabic{figure}}
\renewcommand{\theHequation}{S\arabic{equation}}

\setcounter{dbltopnumber}{3}
\renewcommand{\dbltopfraction}{0.85}
\renewcommand{\dblfloatpagefraction}{0.7}

\maketitlesupplementary

\section{Overview}
\label{sec:suppl_overview}

This supplementary material gives the details that the main paper defers.
Tables, figures, and equations numbered with an S prefix are in this document; those without it are in the main paper.
\Cref{sec:suppl_observations} describes the camera views, the terminal window, and the object reference.
\Cref{sec:suppl_data} describes the models, the three rollout sets, and the human annotation.
\Cref{sec:suppl_prompts} lists the prompts.
\Cref{sec:suppl_conditions} gives the task conditions, the 44 terminal classes, and their severities.
\Cref{sec:suppl_lite} details the PROVE-Lite architecture and training.
\Cref{sec:suppl_eval} reports extended results.
\Cref{sec:suppl_ppo} describes the PPO protocol.
\Cref{sec:suppl_cases} shows qualitative examples.

\section{Observations and Object References}
\label{sec:suppl_observations}

\subsection{Camera Views}
\label{sec:suppl_camera_setup}

Each rollout is recorded by three synchronized cameras: left, right, and wrist (\cref{tab:suppl_camera_views}).
The initial scene $I_{\text{init}}$ contains the three views before execution.
It helps identify the target object and the initial state of relevant fixtures.

\begin{table}[!htb]
\centering
\small
\setlength{\tabcolsep}{4pt}
\begin{tabular}{lll}
\toprule
\textbf{View} & \textbf{Symbol} & \textbf{Role} \\
\midrule
Left camera & $I_{\text{left}}$ & Scene and object context \\
Right camera & $I_{\text{right}}$ & Scene and object context \\
Wrist camera & $I_{\text{wrist}}$ & Close-range manipulation view \\
\bottomrule
\end{tabular}
\caption{Camera views used by PROVE and PROVE-Lite. Camera poses follow the RoboCasa365 rendering configuration.}
\label{tab:suppl_camera_views}
\end{table}

Each camera provides RGB images at $256\times256$ resolution.
The simulator camera identifiers are:
\begin{itemize}
    \setlength{\itemsep}{0pt}
    \setlength{\parsep}{0pt}
    \item Left: {\small\texttt{robot0\_agentview\_left}}
    \item Right: {\small\texttt{robot0\_agentview\_right}}
    \item Wrist: {\small\texttt{robot0\_eye\_in\_hand}}
\end{itemize}

For the Xiaomi robot policy, each view undergoes a 95\% center crop, yielding approximately $243\times243$ pixels, followed by bilinear resizing to $256\times256$.
The three views are supplied separately to the policy.

For PROVE and PROVE-Lite evaluation, no additional manual cropping is applied.
The views are concatenated horizontally in left--right--wrist order before model processing, as detailed in \cref{sec:suppl_terminal_video}.

\begin{figure}[t]
\centering
\includegraphics[width=0.5\textwidth]{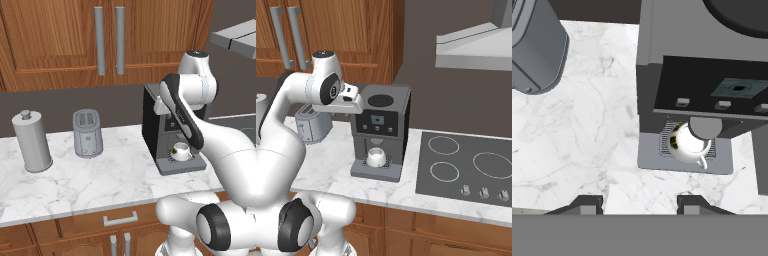}
\caption{\textbf{Initial observation supplied to PROVE.}
Left, right, and wrist views before execution.}
\label{fig:suppl_three_view}
\end{figure}

\subsection{Terminal Window}
\label{sec:suppl_terminal_video}

The terminal observation $I_{\text{final}}$ covers the last second of the episode.
It is sampled at six time steps (\cref{tab:suppl_temporal_sampling}).
At each time step, the synchronized left, right, and wrist views, each at $256\times256$ resolution, are concatenated horizontally in that order to form a single $768\times256$ RGB image.
PROVE and PROVE-Lite both receive the resulting six composite images in chronological order.
No additional manual cropping is applied.
The images are then processed by the model processor configured with \texttt{min\_pixels}$=65{,}536$ and \texttt{max\_pixels}$=196{,}608$.

\begin{table}[!htb]
\centering
\small
\begin{tabular}{cc}
\toprule
\textbf{Frame} & \textbf{Time before termination} \\
\midrule
F1 & $1.0\,\text{s}$ \\
F2 & $0.8\,\text{s}$ \\
F3 & $0.6\,\text{s}$ \\
F4 & $0.4\,\text{s}$ \\
F5 & $0.2\,\text{s}$ \\
F6 & $0.0\,\text{s}$ \\
\bottomrule
\end{tabular}
\caption{Temporal sampling of the terminal window. Each time step contains left, right, and wrist views.}
\label{tab:suppl_temporal_sampling}
\end{table}

\begin{figure*}[!t]
\centering
\begin{minipage}[t]{0.32\textwidth}
\centering
{\small\textbf{F1}\quad $-1.0$ s}\par\smallskip
\includegraphics[width=\linewidth]{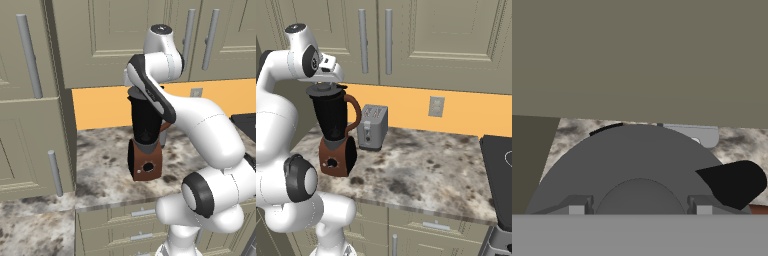}
\end{minipage}\hfill
\begin{minipage}[t]{0.32\textwidth}
\centering
{\small\textbf{F2}\quad $-0.8$ s}\par\smallskip
\includegraphics[width=\linewidth]{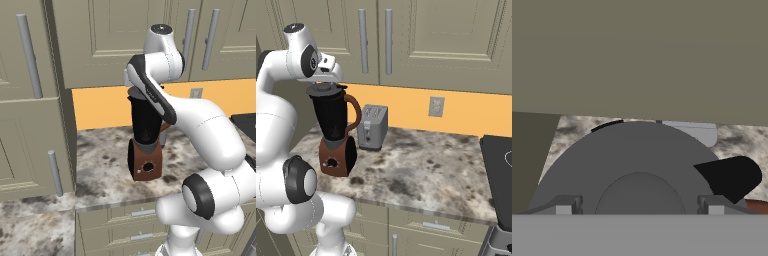}
\end{minipage}\hfill
\begin{minipage}[t]{0.32\textwidth}
\centering
{\small\textbf{F3}\quad $-0.6$ s}\par\smallskip
\includegraphics[width=\linewidth]{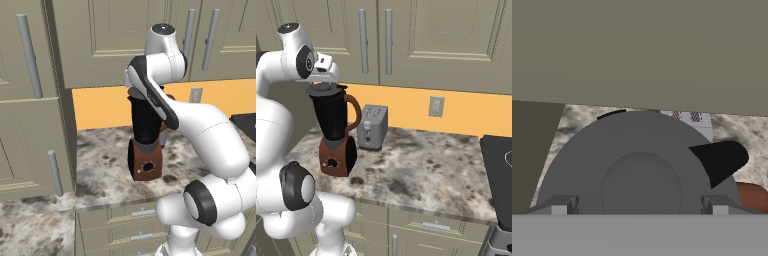}
\end{minipage}
\par\medskip
\begin{minipage}[t]{0.32\textwidth}
\centering
{\small\textbf{F4}\quad $-0.4$ s}\par\smallskip
\includegraphics[width=\linewidth]{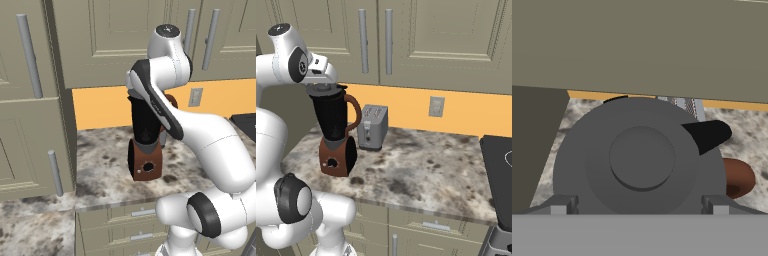}
\end{minipage}\hfill
\begin{minipage}[t]{0.32\textwidth}
\centering
{\small\textbf{F5}\quad $-0.2$ s}\par\smallskip
\includegraphics[width=\linewidth]{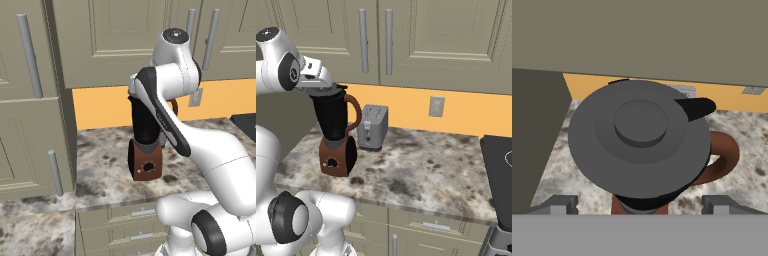}
\end{minipage}\hfill
\begin{minipage}[t]{0.32\textwidth}
\centering
{\small\textbf{F6}\quad $0.0$ s}\par\smallskip
\includegraphics[width=\linewidth]{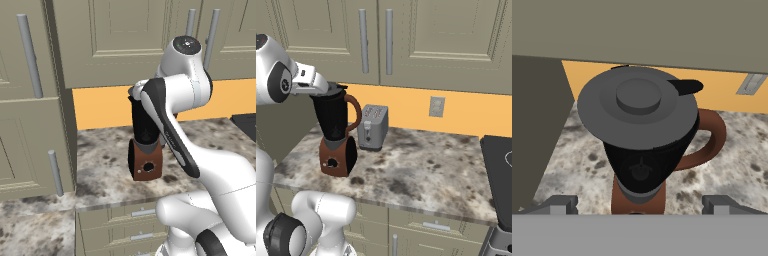}
\end{minipage}
\caption{\textbf{Terminal observations on CloseBlenderLid.}
Six composite images cover the final second of the episode,
ordered chronologically from left to right and top to bottom.
Each image contains synchronized left, right, and wrist views.
Timestamps are relative to episode termination.}
\label{fig:suppl_terminal_window}
\end{figure*}

\subsection{Target-Object Reference}
\label{sec:suppl_object_reference}

The task specification names the target object.
Its simulator object ID identifies the target instance in the initial scene, avoiding confusion among objects of the same category.
The reference $I_{\text{obj}}$ is constructed from target-centered crops of the initial left, right, and wrist views.
When the target is occluded, virtual-camera viewpoints are selected automatically as needed to obtain an unobstructed view of the same object.
The reference is assembled as three horizontally arranged $384\times384$ panels, forming a $1152\times384$ RGB image.
$I_{\text{obj}}$ is used only by M2 and M3; PROVE-Lite does not receive it.

\begin{table}[!htb]
\centering
\small
\setlength{\tabcolsep}{4pt}
\begin{tabular}{@{}>{\raggedright\arraybackslash}p{2.9cm}>{\raggedright\arraybackslash}p{4.9cm}@{}}
\toprule
\textbf{Property} & \textbf{Setting} \\
\midrule
Object identity & Task specification and simulator object ID \\
Reference time & Initial scene \\
Physical cameras & Left, right, wrist \\
Occlusion handling & Virtual-camera rendering of an unobstructed target view \\
Virtual-camera placement & Automatically selected based on target visibility \\
Reference resolution & $1152\times384$ pixels; three horizontally arranged $384\times384$ panels \\
Number of virtual views & As needed to obtain an unobstructed target view \\
\bottomrule
\end{tabular}
\caption{Target-object reference settings.}
\label{tab:suppl_object_reference_settings}
\end{table}

The target-appearance references are constructed from scene RGB renderings conditioned on scene metadata. Simulator segmentation information is used to identify
target regions for cropping.
The reference metadata records the selected view azimuths,
target bounding boxes, and visible-pixel counts.
These references do not reproduce the original scene
poses and are used only for target appearance and
identification, not as evidence of the original initial
or terminal state.

\paragraph{Use of simulator information}
Simulator metadata is used to locate and crop the target
instance for $I_{\text{obj}}$.
During condition verification, $f_{\text{eval}}$ receives
only $\mathcal{I}$, the textual conditions $C_3$, and the
terminal RGB frames, without simulator-state inputs.
PROVE-Lite and its reward computation likewise use no
simulator-state inputs.
On a physical robot, the same appearance reference could
be obtained without privileged state, for example from an
open-vocabulary detector or segmenter applied to the initial
views, or from a photo of the target object.
We have not evaluated this substitution, and we list it
with real-robot validation as future work.

\begin{figure}[t]
\centering
\includegraphics[width=\linewidth]
{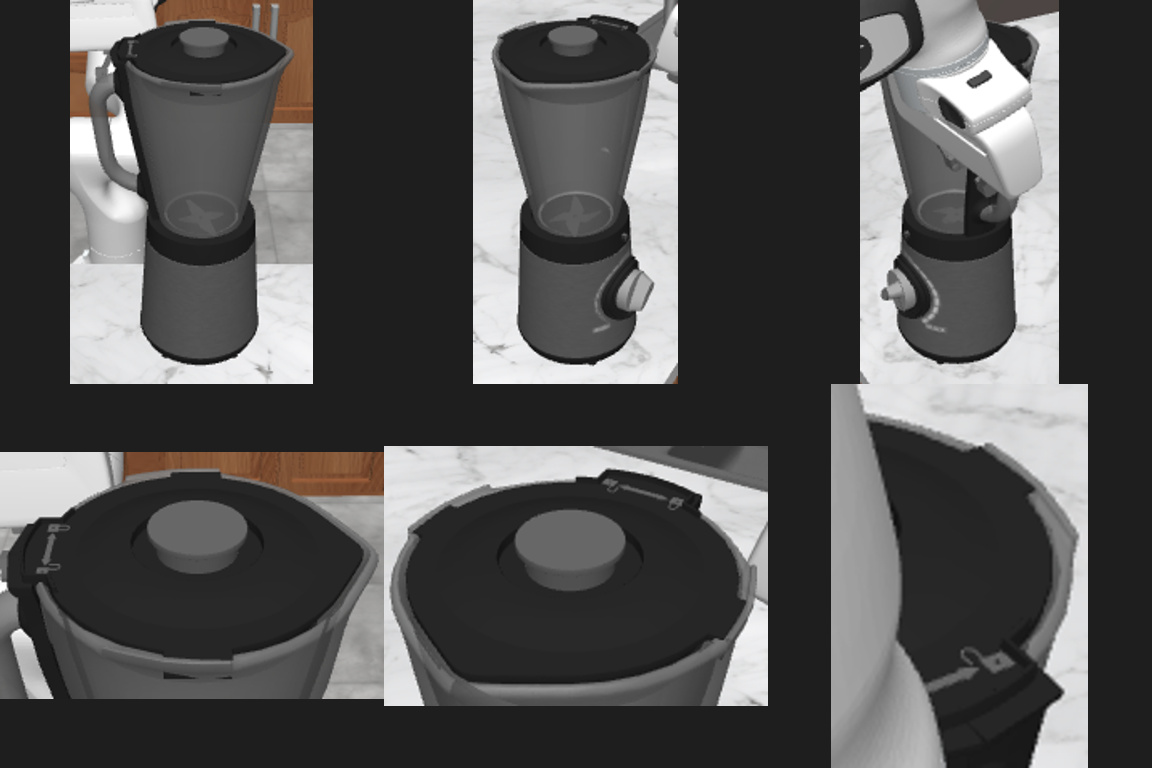}
\caption{\textbf{Target-object reference.}
An example reference image for OpenBlenderLid,
showing the target blender and lid from multiple viewpoints.
The reference is supplied to M2 and M3 for target identification
and condition refinement.}
\label{fig:suppl_object_reference}
\end{figure}

\section{Models, Data, and Annotation}
\label{sec:suppl_data}

\subsection{Models}
\label{sec:suppl_models}

\begin{table*}[t]
\centering
\small
\setlength{\tabcolsep}{4pt}
\newcolumntype{L}[1]{>{\raggedright\arraybackslash}p{#1}}
\begin{tabular*}{\textwidth}{@{\extracolsep{\fill}}L{3.0cm}L{4.4cm}L{3.6cm}L{4.6cm}@{}}
\toprule
\textbf{Component} & \textbf{Model} & \textbf{Configuration} & \textbf{Role} \\
\midrule
PROVE MLLM $\mathcal{M}$ & \texttt{Qwen3-VL-8B-Instruct} & Frozen, BF16 & M1--M3 and verification \\
Goal-image generation & \texttt{gpt-image-2} & Default API settings & Predicted goal image $\hat{I}_{\text{goal}}$ \\
Mapping VLM & \texttt{Qwen3-VL-8B-Instruct} & Frozen, BF16 & Maps PROVE outputs to the condition schema \\
PROVE-Lite backbone & \texttt{Qwen3.5-4B} & $d=2560$; LoRA $r=16$, $\alpha=32$ & Single-pass evaluator \\
\midrule
Generative baselines & \texttt{Qwen3-VL-8B}, \texttt{Qwen3.5-4B} & LoRA $r=16$ & Text-generating comparison \\
Direct judge & GPT-6.1 Sol & Reasoning effort: medium & Direct judge (Tab.~2, row~2); alternative PROVE backbone (row~8) \\
\midrule
Policy & \texttt{RLinf-pi0-robocasa-\allowbreak pretrain-human300} & Policy LoRA $r=16$ & PPO initialization \\
\bottomrule
\end{tabular*}
\caption{Models used in this work, grouped into PROVE and PROVE-Lite components, comparison methods (Tab.~2 and Tab.~4 of the main paper), and the PPO policy.
The generative baselines are LoRA-tuned on the same PROVE pseudo-labels as PROVE-Lite and output the verdict as text.}
\label{tab:suppl_models}
\end{table*}

\paragraph{Software environments}
PROVE and PROVE-Lite
use PyTorch 2.8.0
with CUDA 12.8, Torchvision 0.23.0,
Transformers 5.17.0, and PEFT 0.21.0.
PPO post-training runs in a separate RLinf environment with Transformers 4.53.2,
PEFT 0.17.1, MuJoCo 3.3.1, and Ray 2.48.0.
A separate portable evaluator manifest specifies
Transformers 5.22.1 and PEFT 0.21.2.

\paragraph{Inference settings}
The Qwen3-VL-8B evaluation runner uses BF16
and non-sampling decoding
(\texttt{do\_sample=False}), with a maximum of
650 new tokens per call.
This limit applies to each model call; Table~4 in the
main paper reports the mean number of tokens generated
per rollout.
The generative 4B baseline runner also
uses non-sampling decoding, with a maximum of
256 new tokens.
PROVE-Lite performs one forward pass without
text generation and uses a success threshold of 0.5.

\paragraph{API usage dates}
GPT-6.1 Sol was primarily accessed on
November 15--16, 2026, and \texttt{gpt-image-2}
on November 1--3, 2026.

\subsection{Rollout Sets}
\label{sec:suppl_splits}

We use three rollout sets (\cref{tab:suppl_splits}).
In all sets, devices are selected randomly within the scene setup.

\paragraph{Benchmark}
The benchmark has 6,500 rollouts of the RoboCasa365 Xiaomi-Robotics-1 policy: 65 actions, 10 scenes per action, and 10 rollouts per scene.
Humans label 3,762 successes and 2,738 failures.
The four policies re-evaluated in Sec.~5.3 run on the same 6,500 initializations.

\paragraph{PROVE-Lite training}
PROVE-Lite trains on 32,500 rollouts: 65 actions, 50 scenes per action, and 10 rollouts per scene, with PROVE pseudo-labels only.
These scenes use different layout--style combinations from the benchmark scenes, and the two sets share no scene.

\paragraph{PPO post-training}
Post-training uses 15 actions, each with 80 training and 20 held-out test initializations.
These initializations come from benchmark scenes; training and testing are separated at the initialization level rather than the scene level.
None of these scenes is used to train PROVE-Lite.

\begin{table*}[t]
\centering
\small
\begin{tabular*}{\textwidth}{@{\extracolsep{\fill}}lccccl@{}}
\toprule
\textbf{Set} & \textbf{Actions} & \textbf{Scenes/action} & \textbf{Rollouts/scene} & \textbf{Total} & \textbf{Labels} \\
\midrule
Benchmark & 65 & 10 & 10 & 6,500 & Human (3,762 S / 2,738 F) \\
PROVE-Lite training & 65 & 50 & 10 & 32,500 & PROVE pseudo-labels \\
PPO post-training & 15 & 10 & 8 train + 2 test & 1,200 train + 300 test inits. & Reward model / PROVE \\
\bottomrule
\end{tabular*}
\caption{Rollout sets. Benchmark and PROVE-Lite training scenes are disjoint.}
\label{tab:suppl_splits}
\end{table*}

\subsection{Pseudo-Label Generation}
\label{sec:suppl_pseudolabels}

PROVE labels each of the 32,500 training rollouts.
The binary target is PROVE's decision, $\tilde{S}=\hat{y}$.
The mapping VLM converts $(C_3,\hat{y},\rho)$ and the terminal frames into per-condition targets $(\tilde{s}_k,\tilde{r}_k,\tilde{\mathbf{e}}_k)$ under the schema of \cref{sec:suppl_conditions}.
Any condition not positively confirmed is labeled \texttt{UNMET}.
Human labels are never used as training targets (\cref{tab:suppl_label_provenance}).

\begin{table}[!htb]
\centering
\small
\setlength{\tabcolsep}{5pt}
\begin{tabular}{lll}
\toprule
\textbf{Target} & \textbf{Training} & \textbf{Evaluation} \\
\midrule
Outcome $S$ & PROVE & Human labels \\
Status $s_k$ & PROVE + mapping VLM & PROVE + mapping VLM \\
Class $r_k$ & PROVE + mapping VLM & PROVE + mapping VLM \\
Evidence $\mathbf{e}_k$ & Mapping VLM & Not evaluated \\
\bottomrule
\end{tabular}
\caption{Label provenance for PROVE-Lite.
Condition-level evaluation labels come from the same PROVE and mapping-VLM pipeline applied to the benchmark rollouts, so they measure agreement with the teacher, not with humans.}
\label{tab:suppl_label_provenance}
\end{table}

\subsection{Human Annotation}
\label{sec:suppl_human_annotation}

A primary annotator with relevant experience judged each benchmark rollout from the instruction, the initial scene, and the terminal multi-view frames.
Full replay was available for ambiguous cases. The primary annotator followed a fixed physical-completion rubric and had no access to simulator predicates or model predictions.

\paragraph{Inter-annotator agreement}
A second annotator independently relabeled 1,000 random benchmark rollouts and matched the primary labels on 95.4\% of them (95\% Wilson interval 93.9--96.5\%).
PROVE reaches 96.83\% agreement with the benchmark
reference labels, compared with 95.4\% agreement
between annotators on the relabeled subset.

\paragraph{Blind audit of policy re-evaluation}
For the four policies in Sec.~5.3, annotators reviewed all 8,677 rollouts where native predicates and PROVE disagree.
They saw neither label.
They sided with PROVE in 8,280 cases (95.42\%, 95\% Wilson interval 94.96--95.84\%) and with native predicates in 397.
The rate is stable across policies, from 94.77\% for GR00T N1.5 to 95.87\% for PRTS (\cref{tab:suppl_reeval_decomp}).
For each disagreement, annotators watched the  rollout video and recorded a binary success judgment.

\subsection{Visual Quality Control}
\label{sec:suppl_visual_qc}

Before annotation and evaluation, an automated pass checks rendered frames for duplicated left and right buffers and for high-frequency shader noise.
It flags identical pixels across views and abnormal Laplacian variance.
Flagged episodes are re-rendered from saved MuJoCo states.

We manually inspected all rollout videos and target-object
references for visual quality.
Initial and terminal observations were extracted
from the rollout videos.

\section{Prompts}
\label{sec:suppl_prompts}

\Cref{tab:suppl_prompt_io} summarizes the inputs and outputs of each prompt.

\begin{table*}[t]
\centering
\small
\begin{tabular*}{\textwidth}{@{\extracolsep{\fill}}p{3.4cm}p{6.6cm}p{5.0cm}@{}}
\toprule
\textbf{Stage} & \textbf{Input} & \textbf{Output} \\
\midrule
M1 ($f_{\text{gen}}$) & $\mathcal{I}$ & $C_1$ \\
M2 ($f_{\text{scene}}$) & $\mathcal{I}, C_1, I_{\text{init}}, I_{\text{obj}}$ & $C_2$ \\
Goal generation & $\mathcal{I}, I_{\text{init}}, I_{\text{obj}}$ & $\hat{I}_{\text{goal}}$ \\
M3 ($f_{\text{goal}}$) & $\mathcal{I}, C_2, I_{\text{init}}, I_{\text{obj}}, \hat{I}_{\text{goal}}$ & $C_3$ \\
Verification ($f_{\text{eval}}$) & $\mathcal{I}, C_3, I_{\text{final}}$ & $\hat{y}, \rho$ \\
Mapping VLM & $C_3, \hat{y}, \rho, I_{\text{final}}$, schema of $\tau$ & $\tilde{s}_k, \tilde{r}_k, \tilde{\mathbf{e}}_k$ \\
PROVE-Lite & $\mathcal{I}, I_{\text{final}}$, schema of $\tau$ & $S, \{s_k, r_k, \mathbf{e}_k\}$ \\
Direct judge (GPT-6.1 Sol) & $\mathcal{I}, I_{\text{final}}$ & Binary outcome \\
\bottomrule
\end{tabular*}
\caption{Inputs and outputs of each prompted stage.}
\label{tab:suppl_prompt_io}
\end{table*}

\subsection{Shared Condition Rules}
\label{sec:suppl_shared_prompt_rules}

M1--M3 share the following condition-generation instruction.
They ask for one to four concise, non-redundant, visually verifiable postconditions in JSON:
\begin{promptbox}
\begin{verbatim}
{"conditions":[{"requirement":"..."}]}
\end{verbatim}
\end{promptbox}
The rules forbid constraints the task does not imply, such as visibility requirements, perfect centering, unobservable history, arbitrary arm retreat, and RGB thresholds.
They also state that the generated goal image is fallible guidance, not proof of success.
\begin{promptbox}
You assess physical robot-task completion. Conditions must
describe only instruction-essential physical results, at most
four concise nonredundant requirements. No visibility
requirements, all-camera requirements, perfect centering,
invisible history, arbitrary cleanup, robot retreat, numerical
thresholds inferred from RGB, or future guarantees. Reference
images are not achieved outcome evidence. A WM prediction is
fallible goal guidance, never proof, mandatory appearance
template or truth. Preserve valid earlier conditions; change
only a specific instruction/grounding error. Never alter
conditions to obtain a desired class balance or method ranking.
Generate conditions, NOT an episode success judgement.
Real final evaluation images and labels are withheld.
Return JSON \verb|{"conditions":[{"requirement":"..."}]}|.
If refining, keep previous wording unchanged unless the
additional allowed input identifies a concrete needed correction.
Do not invent new requirements merely because more input is present.
\end{promptbox}

\subsection{M1: Instruction-Derived Conditions}
M1 receives only the instruction $\mathcal{I}$.
It lists the physical postconditions that must hold at the end of the episode for $\mathcal{I}$ to be satisfied.
Without scene context, it may add plausible but unrequired conditions, such as closing a door that was already open.
The M1 prompt is summarized below.

M1 sends the shared instruction and the applicable task rules
in a single user message, followed by the task name and
instruction. This stage has no separate system message.
\begin{promptbox}
\texttt{[shared instruction]}\\
\texttt{[applicable task rules]}\\
\texttt{Task: [task name]}\\
\texttt{Instruction: [task instruction]}
\end{promptbox}

\subsection{M2: Scene-Grounded Refinement}
M2 receives $\mathcal{I}$, $C_1$, the initial views $I_{\text{init}}$, and the object reference $I_{\text{obj}}$.
It identifies the target instance, removes conditions that the initial scene already satisfies or that the instruction does not require, and makes object references specific.
It does not see any post-action image.
The M2 prompt is summarized below.

M2 prepends the initial scene and target-object reference to
the M1 message and appends $C_1$ as JSON. The image preamble
is translated from Chinese.
\begin{promptbox}
Initial scene and object reference, not evidence of completion:
\texttt{[initial image] [object reference]}\\
\texttt{[M1 message]}\\
\texttt{previous\_conditions:} $[C_1 \text{ as JSON}]$
\end{promptbox}

\subsection{Goal-Image Generation}

\texttt{gpt-image-2} edits the initial scene to depict one
plausible post-action configuration for the task instruction
$\mathcal{I}$. The generation request asks the model to preserve
the target object's identity, camera viewpoints, panel order,
and unrelated scene elements.

In the standard M1--M3 pipeline, M3 uses the generated goal image
$\hat{I}_{\text{goal}}$ to refine the success conditions $C_2$.
Terminal verification checks the resulting conditions against
the recorded terminal observations. The ablations in Tab.~2
also supply the generated goal image during verification.

The generation request is summarized below.

\begin{promptbox}
Edit the initial scene to show a plausible successful outcome
for the task instruction. Apply the changes required by the
instruction while preserving the target object's identity,
camera viewpoints, panel order, and unrelated scene elements.
Keep the resulting scene consistent across the three views.

\medskip
\textbf{Task instruction.} \texttt{[instruction]}
\end{promptbox}

\subsection{M3: Goal-Guided Refinement}
M3 receives $\mathcal{I}$, $C_2$, $I_{\text{init}}$, $I_{\text{obj}}$, and $\hat{I}_{\text{goal}}$.
It uses the goal image to sharpen spatial relations, placement regions, and support requirements, for example by adding that an object must be fully inside, supported, and released.
The prompt states that $\hat{I}_{\text{goal}}$ shows one possible outcome, may contain artifacts, and must not be treated as evidence.
Only geometric and relational cues enter the textual conditions $C_3$.
The M3 prompt is summarized below.

M3 places the generated goal image before the initial scene
and target-object reference, and supplies $C_2$ as the previous
conditions. The image preambles are translated from Chinese.
\begin{promptbox}
The WM prediction is a fallible goal reference:
\texttt{[generated goal image]}\\
Initial scene and object reference, not evidence of completion:
\texttt{[initial image] [object reference]}\\
\texttt{[M1 message]}\\
\texttt{previous\_conditions:} $[C_2 \text{ as JSON}]$
\end{promptbox}

\subsection{Terminal Verification}
$f_{\text{eval}}$ receives $\mathcal{I}$, the fixed conditions $C_3$, and the terminal observation, but not $\hat{I}_{\text{goal}}$.
The 18 view images (six timestamps $\times$ three camera views) are organized as six composite input images, each horizontally concatenating the synchronized left, right, and wrist views in that order.
Every condition must be positively confirmed; partial cues are not enough.
Multiple views help avoid mistaking occlusion by a nearby
gripper for incomplete release.
A temporal rule uses later frames for transient events such as settling or appliance activation.
On failure, $\rho$ names the primary observable deficiency without speculating about earlier events.
\begin{promptbox}
Verify every fixed condition against real terminal evidence only.
Unconfirmed conditions mean FAILURE. Describe the primary visible
deficiency; do not invent execution history. Return JSON
\verb+{"label":"SUCCESS|FAILURE","reason":"..."}+.
\end{promptbox}

The user message appends a JSON object with keys
\texttt{instruction} and \texttt{fixed\_conditions},
followed by the terminal views.
Each view is preceded by the line\newline
{\footnotesize\verb|Observed F{i} {view} {relative_seconds:+.3f}s|}.
The response is a JSON object with \texttt{label}
(\texttt{SUCCESS} or \texttt{FAILURE}) and
\texttt{reason} (a string).

\subsection{Mapping VLM}
The mapping VLM receives $C_3$, $\hat{y}$, $\rho$, the terminal frames, and the condition schema of action $\tau$.
For each condition in $\mathcal{K}_\tau$, it returns a status, a class from the valid classes of that condition, and the supporting frames and views.
A condition is \texttt{SATISFIED} only if the evidence positively confirms it; otherwise it is \texttt{UNMET}.
The binary target stays PROVE's verdict; the mapping only derives the per-condition diagnosis.
The output must report each active condition exactly once:
\begin{promptbox}
\begin{verbatim}
{"checks":[{"condition_code":...,
  "status":"SATISFIED"|"UNMET",
  "observation_code":...,
  "evidence":[{"frame":6,"view":"left"}],
  "visual_note":...}]}
\end{verbatim}
\end{promptbox}
An output is rejected if a condition is missing or repeated, a class is not valid for its condition, a frame or view is out of range, or a \texttt{SATISFIED} condition has no evidence.
Failed mapping attempts are retried up to three times.
All rollouts ultimately yield valid structured
annotations, giving a final mapping failure rate of 0\%.

\subsection{PROVE-Lite Input Template}
\label{sec:suppl_lite_prompt}
The user turn reads as follows; the chat template adds the role tokens.
\begin{promptbox}
Evaluate the instruction using only real terminal observations. Inspect the task-specific condition schema. Determine overall success and the terminal state of each active condition. Do not infer unobserved history.\\
\texttt{\{"instruction": ..., "condition\_schema": \{"goal": ..., "conditions": \{...\}\}\}}\\
Six composite images in chronological order, containing 18 view images in total; each composite arranges the left, right, and wrist views horizontally in that order.\\
Is the task successfully completed? Inspect the active conditions.
\end{promptbox}
The prompt lists the conditions of the active action
and all 44 terminal classes.
For each active condition, the schema specifies its
rule, definition, and valid classes.
The hidden state at the last input position feeds both heads.

\subsection{Direct-Judge Baseline}
GPT-6.1 Sol receives $\mathcal{I}$ and the terminal observation, comprising 18 view images from six timestamps and three camera views.
The views at each timestamp are concatenated horizontally in left--right--wrist order, yielding six composite images supplied in chronological order.
The model answers whether the task succeeded.
It receives no initial scene, object reference, goal image, or condition list.
The prompt is given below in English translation.
Placeholders are filled with the task name, instruction,
and terminal-image metadata for each rollout.

\begin{promptbox}
You are a visual evaluator of robot-task completion.
Determine whether the current trajectory succeeds or
fails, and explain the decision using concrete visual
evidence. No labeled examples are provided; only the
real terminal images constitute completion evidence.

Success must be supported by the required physical
relation or state in the real images. Do not invent
closure, correct placement, or button activation when
the target is outside the field of view or fully
occluded. Partial visibility may support reasonable
inference across views; a complete object contour is
not required. The absence of an initial image or the
full action history is not itself evidence of failure.

For placement tasks, assess the task-specific location,
orientation, and containment requirements. Failure due
to continued grasping requires clear evidence that the
closed gripper actually holds or supports the object.
Proximity, contact, occlusion, or not observing release
is insufficient. Do not require the arm to retreat.

F1--F6 are terminal observations; F1 is not the initial
state. Judge only the available views and frames.
Do not invent missing observations.

Return only the required JSON. Do not use tools,
browse, read other files, or output a reasoning trace.
If the evidence is insufficient to confirm success,
return FAILURE and set \texttt{evidence\_insufficient}
to true.
The label and explanation must express the same conclusion.

Task: \texttt{[task name]}.\\
Instruction: \texttt{[episode instruction]}.\\
Attachments: real F1--F6, each identified by its
simulation step and time relative to termination.\\
Camera order: left, right, wrist.
\end{promptbox}

The structured response contains
\texttt{label},
\texttt{reason},
\texttt{target\_visible},
\texttt{failure\_basis},
\texttt{grasp\_status},
\texttt{grasp\_evidence}, and
\texttt{evidence\_insufficient}.
The label is either \texttt{SUCCESS} or \texttt{FAILURE};
the explanation is returned in Chinese.

\section{Task Conditions, Terminal Classes, and Severities}
\label{sec:suppl_conditions}

\subsection{Conditions and the 44 Terminal Classes}
\label{sec:suppl_44dict}

The schema has 10 generic conditions (\cref{tab:suppl_semantics}).
Each condition keeps the identifier of the action family it was first written for, but its meaning is generic.
These identifiers are used consistently in the schemas
provided to the mapping VLM and PROVE-Lite.
For example, \texttt{rack\_target} covers any part, setting, or robot pose that must reach an instructed target, and \texttt{kettle\_active} covers any device that must reach an instructed on, off, or running state.

\begin{table}[!htb]
\centering
\footnotesize
\setlength{\tabcolsep}{3pt}
\begin{tabular}{@{}lp{5.2cm}@{}}
\toprule
\textbf{Condition} & \textbf{Meaning} \\
\midrule
\texttt{door\_closed} & A door, drawer, or appliance head is fully closed \\
\texttt{lid\_seated} & A lid is seated on its opening \\
\texttt{mug\_aligned} & The mug is under the dispenser \\
\texttt{object\_inside} & The object is fully inside the instructed container \\
\texttt{object\_counter} & The object is on the instructed surface or region \\
\texttt{stable\_placement} & The object is released and rests stably \\
\texttt{rack\_target} & A part, setting, or robot pose reaches the instructed target \\
\texttt{kettle\_active} & A device reaches the instructed on, off, or running state \\
\texttt{microwave\_active} & An appliance reaches the instructed running state \\
\texttt{spout\_target} & The faucet spout reaches the instructed direction \\
\bottomrule
\end{tabular}
\caption{The 10 generic task conditions.}
\label{tab:suppl_semantics}
\end{table}

\Cref{tab:suppl_44dict} lists all 44 terminal classes.
Each class belongs to one condition and has a fixed penalty $d(r)$.
A class with $d(r)=0$ has status \texttt{SATISFIED}; every other class, including \texttt{unknown}, has status \texttt{UNMET}.
The tiers match Tab.~1: 12 classes at $d=0$, 2 at $0.3$, 4 at $0.6$, 2 at $0.8$, and 24 at $1.0$.
The same table supervises the ConditionHead and computes $C$ for the reward.

\begin{table*}[t]
\centering
\footnotesize
\setlength{\tabcolsep}{4pt}
\renewcommand{\arraystretch}{0.92}
\begin{tabular*}{\textwidth}{@{\extracolsep{\fill}}rllp{7.4cm}cc@{}}
\toprule
\textbf{\#} & \textbf{Condition} & \textbf{Class $r$} & \textbf{Observable terminal state} & \textbf{Status} & \textbf{$d(r)$} \\
\midrule
1 & \texttt{door\_closed} & \texttt{closed} & Flush against the body & SAT. & 0.0 \\
2 &  & \texttt{gap} & A visible gap remains & UNMET & 0.3 \\
3 &  & \texttt{open} & Still clearly open & UNMET & 1.0 \\
4 &  & \texttt{unknown} & Cannot be confirmed from the frames & UNMET & 1.0 \\
\midrule
5 & \texttt{lid\_seated} & \texttt{seated} & Aligned with the opening and closed & SAT. & 0.0 \\
6 &  & \texttt{offset} & Off the opening or tilted & UNMET & 0.3 \\
7 &  & \texttt{open} & Opening still exposed & UNMET & 1.0 \\
8 &  & \texttt{unknown} & Cannot be confirmed from the frames & UNMET & 1.0 \\
\midrule
9 & \texttt{mug\_aligned} & \texttt{aligned} & Mug opening under the nozzle & SAT. & 0.0 \\
10 &  & \texttt{offset} & Mug off the nozzle & UNMET & 0.6 \\
11 &  & \texttt{absent} & No target mug under the nozzle & UNMET & 1.0 \\
12 &  & \texttt{unknown} & Cannot be confirmed from the frames & UNMET & 1.0 \\
\midrule
13 & \texttt{object\_inside} & \texttt{inside} & Fully inside the target region & SAT. & 0.0 \\
14 &  & \texttt{boundary} & Straddles the boundary & UNMET & 0.6 \\
15 &  & \texttt{outside} & Not in the target region & UNMET & 1.0 \\
16 &  & \texttt{unknown} & Cannot be confirmed from the frames & UNMET & 1.0 \\
\midrule
17 & \texttt{object\_counter} & \texttt{supported} & Supported at the target & SAT. & 0.0 \\
18 &  & \texttt{held} & Still supported by the gripper & UNMET & 0.8 \\
19 &  & \texttt{off\_counter} & Away from the target & UNMET & 1.0 \\
20 &  & \texttt{unknown} & Cannot be confirmed from the frames & UNMET & 1.0 \\
\midrule
21 & \texttt{stable\_placement} & \texttt{supported} & Rests stably on the target surface & SAT. & 0.0 \\
22 &  & \texttt{held} & Still grasped & UNMET & 0.8 \\
23 &  & \texttt{falling} & Visibly tips or falls & UNMET & 1.0 \\
24 &  & \texttt{uncertain\_support} & Support occluded or ambiguous & UNMET & 1.0 \\
25 &  & \texttt{off\_counter} & Away from the required location & UNMET & 1.0 \\
26 &  & \texttt{unknown} & Cannot be confirmed from the frames & UNMET & 1.0 \\
\midrule
27 & \texttt{rack\_target} & \texttt{reached} & At the target & SAT. & 0.0 \\
28 &  & \texttt{partial} & Progress, but short of the target & UNMET & 0.6 \\
29 &  & \texttt{wrong\_direction} & No correct progress, or moved the wrong way & UNMET & 1.0 \\
30 &  & \texttt{unknown} & Cannot be confirmed from the frames & UNMET & 1.0 \\
\midrule
31 & \texttt{kettle\_active} & \texttt{indicator\_on} & Direct state evidence, e.g., indicator, flame, or water flow & SAT. & 0.0 \\
32 &  & \texttt{switch\_active} & Reliable evidence from the control, e.g., a switch or knob & SAT. & 0.0 \\
33 &  & \texttt{visible\_off} & Clearly in the opposite state & UNMET & 1.0 \\
34 &  & \texttt{no\_visible\_activation} & No evidence of the required state & UNMET & 1.0 \\
35 &  & \texttt{unknown} & Cannot be confirmed from the frames & UNMET & 1.0 \\
\midrule
36 & \texttt{microwave\_active} & \texttt{display\_running} & Display supports the target state & SAT. & 0.0 \\
37 &  & \texttt{interior\_active} & Interior or other direct cue supports it & SAT. & 0.0 \\
38 &  & \texttt{visible\_off} & Evidence of the opposite state & UNMET & 1.0 \\
39 &  & \texttt{no\_visible\_activation} & No evidence of the required state & UNMET & 1.0 \\
40 &  & \texttt{unknown} & Cannot be confirmed from the frames & UNMET & 1.0 \\
\midrule
41 & \texttt{spout\_target} & \texttt{reached} & At the instructed direction & SAT. & 0.0 \\
42 &  & \texttt{partial} & Same direction, short of the target & UNMET & 0.6 \\
43 &  & \texttt{wrong\_direction} & Wrong direction or no correct progress & UNMET & 1.0 \\
44 &  & \texttt{unknown} & Cannot be confirmed from the frames & UNMET & 1.0 \\
\bottomrule
\end{tabular*}
\caption{All 44 terminal classes with their status and penalty. SAT.\ denotes \texttt{SATISFIED}. Only the classes of an active condition are valid for that condition.}
\label{tab:suppl_44dict}
\end{table*}

\subsection{Per-Action Conditions and Weights}
\label{sec:suppl_task_weights}

For action $\tau$ with conditions $\mathcal{K}_\tau$, the completion score is $C=1-\sum_{k}w_k d(r_k)$ with $\sum_k w_k=1$.
\Cref{tab:suppl_65} groups all 65 benchmark actions by their conditions and weights.
Placement actions put 0.7 on reaching the target and 0.3 on stable release.
All other actions except AdjustWaterTemperature and CoffeeSetupMug use a single condition.
\Cref{tab:suppl_ppo_weights} lists the 15 PPO actions.

\begin{table*}[t]
\centering
\footnotesize
\setlength{\tabcolsep}{4pt}
\begin{tabular*}{\textwidth}{@{\extracolsep{\fill}}>{\raggedright\arraybackslash}p{3.4cm}cp{12.4cm}@{}}
\toprule
\textbf{Conditions and weights $w_k$} & \textbf{\#} & \textbf{Actions} \\
\midrule
\texttt{rack\_target} 1.0 & 18 & AdjustToasterOvenTemperature, LowerHeat, NavigateKitchen, OpenBlenderLid, OpenCabinet, OpenDishwasher, OpenDrawer, OpenElectricKettleLid, OpenFridge, OpenFridgeDrawer, OpenMicrowave, OpenOven, OpenStandMixerHead, OpenToasterOvenDoor, PreheatOven, SlideDishwasherRack, SlideOvenRack, SlideToasterOvenRack \\
\addlinespace[2pt]
\texttt{object\_counter} 0.7; \texttt{stable\_placement} 0.3 & 12 & CheesyBread, CoffeeServeMug, PickPlaceCabinetToCounter, PickPlaceCounterToSink, PickPlaceCounterToStove, PickPlaceDrawerToCounter, PickPlaceFridgeDrawerToShelf, PickPlaceMicrowaveToCounter, PickPlaceSinkToCounter, PickPlaceStoveToCounter, PickPlaceToasterOvenToCounter, PickPlaceToasterToCounter \\
\addlinespace[2pt]
\texttt{object\_inside} 0.7; \texttt{stable\_placement} 0.3 & 10 & MakeIcedCoffee, PackDessert, PickPlaceCounterToBlender, PickPlaceCounterToCabinet, PickPlaceCounterToDrawer, PickPlaceCounterToMicrowave, PickPlaceCounterToOven, PickPlaceCounterToStandMixer, PickPlaceCounterToToasterOven, PickPlaceFridgeShelfToDrawer \\
\addlinespace[2pt]
\texttt{door\_closed} 1.0 & 9 & CloseCabinet, CloseDishwasher, CloseDrawer, CloseFridge, CloseFridgeDrawer, CloseMicrowave, CloseOven, CloseStandMixerHead, CloseToasterOvenDoor \\
\addlinespace[2pt]
\texttt{kettle\_active} 1.0 & 9 & StartCoffeeMachine, TurnOffSinkFaucet, TurnOffStove, TurnOnBlender, TurnOnElectricKettle, TurnOnSinkFaucet, TurnOnStove, TurnOnToaster, TurnOnToasterOven \\
\addlinespace[2pt]
\texttt{lid\_seated} 1.0 & 2 & CloseBlenderLid, CloseElectricKettleLid \\
\addlinespace[2pt]
\texttt{microwave\_active} 1.0 & 2 & TurnOffMicrowave, TurnOnMicrowave \\
\addlinespace[2pt]
\texttt{rack\_target} 0.7; \texttt{kettle\_active} 0.3 & 1 & AdjustWaterTemperature \\
\addlinespace[2pt]
\texttt{mug\_aligned} 0.7; \texttt{stable\_placement} 0.3 & 1 & CoffeeSetupMug \\
\addlinespace[2pt]
\texttt{spout\_target} 1.0 & 1 & TurnSinkSpout \\
\bottomrule
\end{tabular*}
\caption{Conditions and weights of all 65 benchmark actions, grouped by condition set.}
\label{tab:suppl_65}
\end{table*}

\begin{table}[!htb]
\centering
\footnotesize
\setlength{\tabcolsep}{2.5pt}
\begin{tabular}{@{}l>{\raggedright\arraybackslash}p{3.05cm}r@{}}
\toprule
\textbf{Action} & \textbf{Conditions and weights} & \textbf{Steps} \\
\midrule
CoffeeSetupMug & \texttt{mug\_aligned} 0.7;\newline \texttt{stable\_placement} 0.3 & 600 \\
TurnSinkSpout & \texttt{spout\_target} 1.0 & 300 \\
CloseDishwasher & \texttt{door\_closed} 1.0 & 300 \\
SlideDishwasherRack & \texttt{rack\_target} 1.0 & 450 \\
TurnOnMicrowave & \texttt{microwave\_active} 1.0 & 450 \\
TurnOnElectricKettle & \texttt{kettle\_active} 1.0 & 450 \\
PickPlaceCounterToCabinet & \texttt{object\_inside} 0.7;\newline \texttt{stable\_placement} 0.3 & 750 \\
CloseBlenderLid & \texttt{lid\_seated} 1.0 & 900 \\
CloseFridge & \texttt{door\_closed} 1.0 & 900 \\
PickPlaceSinkToCounter & \texttt{object\_counter} 0.7;\newline \texttt{stable\_placement} 0.3 & 900 \\
PickPlaceCabinetToCounter & \texttt{object\_counter} 0.7;\newline \texttt{stable\_placement} 0.3 & 450 \\
PickPlaceStoveToCounter & \texttt{object\_counter} 0.7;\newline \texttt{stable\_placement} 0.3 & 450 \\
PickPlaceToasterOvenToCounter & \texttt{object\_counter} 0.7;\newline \texttt{stable\_placement} 0.3 & 450 \\
PickPlaceToasterToCounter & \texttt{object\_counter} 0.7;\newline \texttt{stable\_placement} 0.3 & 600 \\
PickPlaceMicrowaveToCounter & \texttt{object\_counter} 0.7;\newline \texttt{stable\_placement} 0.3 & 750 \\
\bottomrule
\end{tabular}
\caption{Conditions, weights, and episode horizons of the 15 PPO actions.}
\label{tab:suppl_ppo_weights}
\end{table}

\subsection{Reward Walkthrough}
\label{sec:suppl_reward_walkthrough}

For PickPlaceCounterToCabinet, $\mathcal{K}_\tau=\{\texttt{object\_inside}, \texttt{stable\_placement}\}$ with weights 0.70 and 0.30.
Suppose the object rests stably on the shelf but straddles the cabinet rim.
Then $r_1=\texttt{boundary}$ with $d=0.6$ and $r_2=\texttt{supported}$ with $d=0$, so
\begin{equation}
    C = 1 - (0.70\times0.6 + 0.30\times0) = 0.58.
\end{equation}
With $S=0$ and $\lambda=0.8$, the reward is $R=0.2\times0.58=0.116$.
\Cref{tab:suppl_fridge_reward} gives the single-condition CloseFridge case.

\begin{table}[!htb]
\centering
\small
\begin{tabular}{lcccc}
\toprule
\textbf{Terminal state} & $d$ & $S$ & $C$ & $R$ \\
\midrule
Wide-open door & 1.0 & 0 & 0.0 & 0.00 \\
Narrow gap & 0.3 & 0 & 0.7 & 0.14 \\
Fully closed & 0.0 & 1 & 1.0 & 1.00 \\
\bottomrule
\end{tabular}
\caption{CloseFridge reward with $w=1$ and $\lambda=0.8$.}
\label{tab:suppl_fridge_reward}
\end{table}

\begin{figure*}[t]
\centering
\begin{minipage}[t]{0.49\textwidth}
\centering
{\small\textbf{Earlier}}\par\smallskip
\includegraphics[width=\linewidth]
{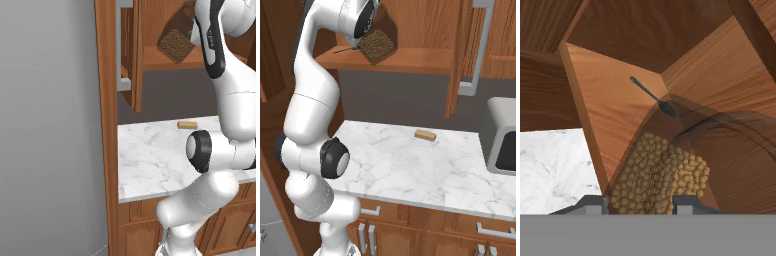}
\end{minipage}\hfill
\begin{minipage}[t]{0.49\textwidth}
\centering
{\small\textbf{Final}}\par\smallskip
\includegraphics[width=\linewidth]
{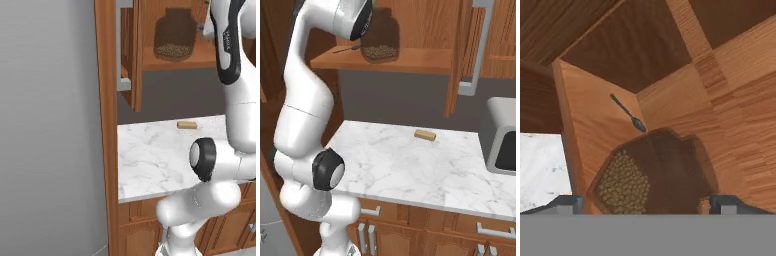}
\end{minipage}
\caption{\textbf{Multi-condition reward example.}
Earlier and final observations from the same
\texttt{PickPlaceCounterToCabinet} rollout.
In this manually inspected example, the object straddles
the cabinet rim but is supported.
The containment condition is unmet
($r_1=\texttt{boundary}$, $d_1=0.6$),
while stable placement is satisfied
($r_2=\texttt{supported}$, $d_2=0$).
With weights $w_1=0.7$ and $w_2=0.3$,
$C=0.7(1-0.6)+0.3(1-0)=0.58$.
For $S=0$ and $\lambda=0.8$, the resulting reward is
$R=0.8S+0.2C=0.116$.}
\label{fig:suppl_reward_example}
\end{figure*}

\section{PROVE-Lite Architecture and Training}
\label{sec:suppl_lite}

\subsection{Input}
\label{sec:suppl_lite_input}

The input is one user turn.
It states the task, then gives the instruction,
the conditions of the active action, and all
44 terminal classes in JSON format.
The schema specifies the action goal and, for each
active condition, its rule, definition, and valid classes.
The terminal observation contains 18 view images from six timestamps and three camera views.
At each timestamp, the synchronized left, right, and wrist views are concatenated horizontally in that order, yielding six composite images supplied in chronological order.
The turn ends with the query ``Is the task successfully completed? Inspect the active conditions.''
The tokenized input contains exactly six composite images and at most 8,192 tokens.
The auxiliary evidence vector retains 18 positions, one for each timestamp--view pair.
\Cref{sec:suppl_lite_prompt} gives the text.

\subsection{Heads}
\label{sec:suppl_lite_heads}

PROVE-Lite reads the final-layer hidden state $\mathbf{h}\in\mathbb{R}^{2560}$ at the last input position.
Two heads read from $\mathbf{h}$ (Sec.~4.1).
\begin{itemize}
    \item \textbf{DecisionHead} is a linear layer $\mathbb{R}^{2560}\!\to\!\mathbb{R}^{2}$. Its two rows are copied from the \texttt{lm\_head} rows of the single tokens \texttt{false} and \texttt{true}. At initialization, it reproduces the base model's next-token scores restricted to these two tokens. The \texttt{lm\_head} is then removed.
    \item \textbf{ConditionHead} is LayerNorm, Linear(2560, 512), GELU, Dropout(0.1), and Linear(512, $10\times64$). For each of the 10 conditions, the 64 outputs hold 2 status logits, 44 class logits, and 18 evidence logits.
\end{itemize}
Only the conditions of the active action are read and trained.
Class logits outside the valid classes of a condition are masked before the softmax.

\begin{figure*}[!tp]
\centering
\includegraphics[width=\textwidth]{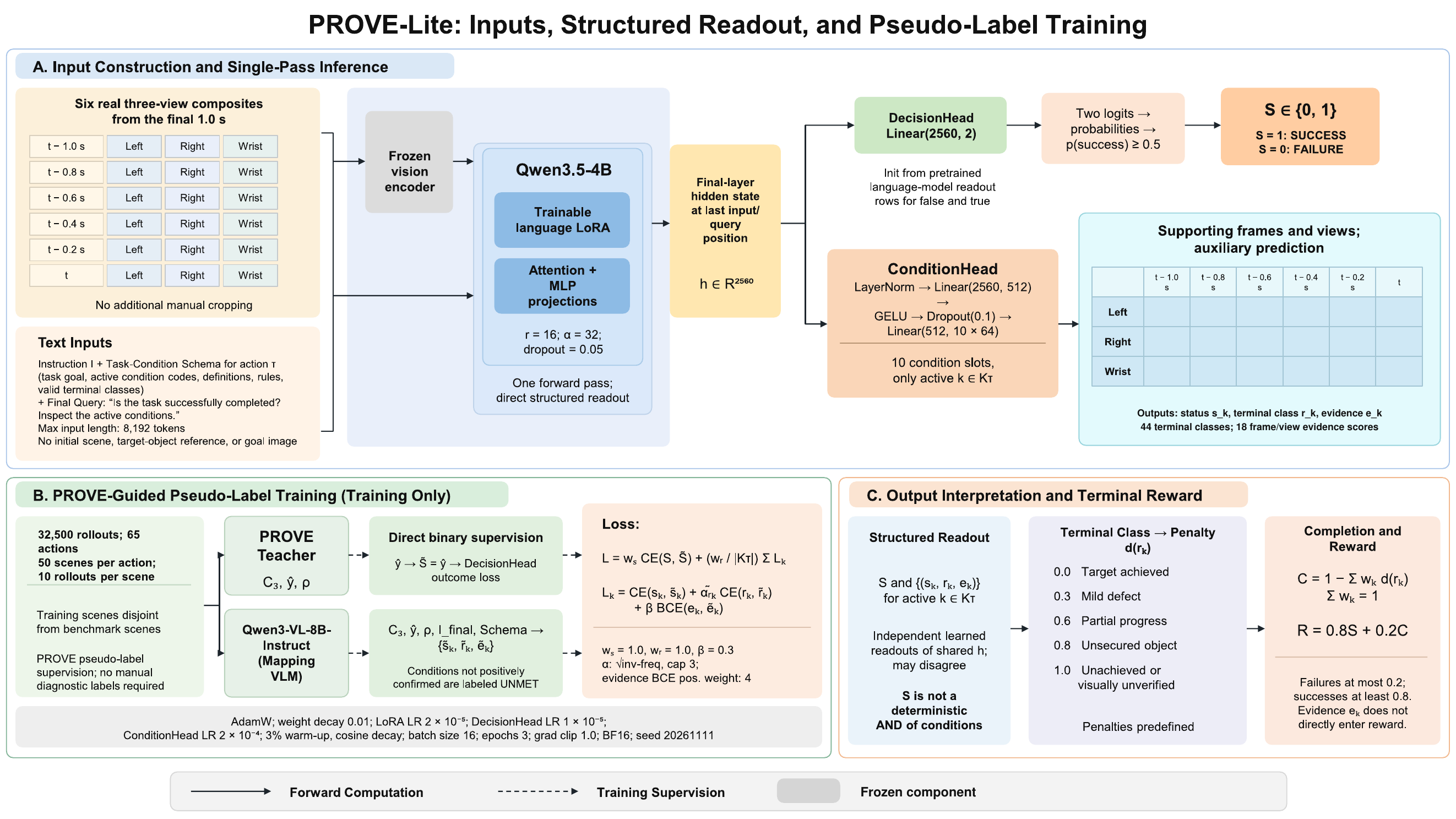}
\caption{\textbf{PROVE-Lite architecture and training.}
A shared query-position feature feeds the DecisionHead
and ConditionHead for binary outcome prediction and
structured condition diagnosis.
PROVE-derived pseudo-labels supervise joint training,
and predicted terminal classes determine the graded reward.}
\label{fig:suppl_lite_arch}
\end{figure*}

\subsection{Training Configuration}
\label{sec:suppl_lite_hparams}

All trainable parameters are optimized in one stage with Eqs.~(6) and~(7).
$\mathcal{L}_k$ is averaged over the active conditions of each rollout.
The class weight $\alpha_r$ is the square root of the inverse class frequency among the valid classes of a condition, capped at 3 and normalized to mean one.
The evidence term uses a positive-class weight of 4.
\Cref{tab:suppl_lite_hparams} lists the configuration.

\begin{table}[!htb]
\centering
\footnotesize
\setlength{\tabcolsep}{3pt}
\begin{tabular}{@{}lp{4.6cm}@{}}
\toprule
\textbf{Parameter} & \textbf{Value} \\
\midrule
Backbone & \texttt{Qwen3.5-4B}, $d=2560$, BF16 \\
Trainable & Language LoRA and both heads \\
Frozen & Vision encoder \\
LoRA & $r=16$, $\alpha=32$, dropout 0.05 \\
LoRA targets & Attention and MLP projections \\
Loss weights & $w_s=1.0$, $w_r=1.0$, $\beta=0.3$ \\
Class weight cap & 3 \\
Optimizer & AdamW, weight decay 0.01 \\
Learning rate & LoRA $2\times10^{-5}$; DecisionHead $1\times10^{-5}$; ConditionHead $2\times10^{-4}$ \\
Schedule & 3\% linear warm-up, cosine decay \\
Batch size, epochs & 16, 3 \\
Gradient clipping & 1.0 \\
Seed & 20261111 \\
Success threshold & 0.5 \\
\bottomrule
\end{tabular}
\caption{PROVE-Lite training configuration.}
\label{tab:suppl_lite_hparams}
\end{table}

PROVE-Lite is trained in parallel on eight
NVIDIA RTX 6000 Ada Generation GPUs.

\section{Extended Results}
\label{sec:suppl_eval}

\subsection{Confusion Counts and Chance-Corrected Agreement}
\label{sec:suppl_confusion}

\Cref{tab:suppl_confusion} expands Tab.~2 into counts on the 3,762 human successes and 2,738 human failures.
It also reports balanced accuracy and Cohen's $\kappa$ with the human labels, which correct for the 58:42 class ratio.
The ranking of evaluators is the same under all three measures.

\begin{table*}[t]
\centering
\small
\setlength{\tabcolsep}{0pt}
\begin{tabular*}{\textwidth}{@{\extracolsep{\fill}}clrrrrccc@{}}
\toprule
\textbf{\#} & \textbf{Evaluator} & \textbf{TP} & \textbf{TN} & \textbf{FP} & \textbf{FN} & \textbf{Agree.\ (95\% CI)} & \textbf{Bal.\ acc.} & \textbf{$\kappa$} \\
\midrule
1 & Simulator predicate & 2,364 & 2,127 & 611 & 1,398 & 69.09 (67.96--70.20) & 70.26 & 0.390 \\
2 & GPT-6.1 Sol, direct judge & 3,257 & 2,341 & 397 & 505 & 86.12 (85.26--86.94) & 86.04 & 0.717 \\
3 & M1 & 3,092 & 2,272 & 466 & 670 & 82.52 (81.58--83.43) & 82.59 & 0.645 \\
4 & M2 & 3,468 & 2,245 & 493 & 294 & 87.89 (87.08--88.66) & 87.09 & 0.749 \\
5 & M3 (PROVE) & 3,716 & 2,578 & 160 & 46 & 96.83 (96.38--97.23) & 96.47 & 0.935 \\
6 & M2 + goal at verification & 3,056 & 2,356 & 382 & 706 & 83.26 (82.33--84.15) & 83.64 & 0.662 \\
7 & M3 + goal at verification & 3,502 & 2,375 & 363 & 260 & 90.42 (89.68--91.11) & 89.92 & 0.802 \\
8 & PROVE, GPT-6.1 Sol backbone & 3,718 & 2,579 & 159 & 44 & 96.88 (96.43--97.27) & 96.51 & 0.936 \\
9 & PROVE-Lite & 3,360 & 2,568 & 170 & 402 & 91.20 (90.49--91.86) & 91.55 & 0.822 \\
\bottomrule
\end{tabular*}
\caption{Confusion counts against human labels on the 6,500 benchmark rollouts. Success is the positive class. The rates in Table~2 of the main paper are computed
from these counts; intervals are 95\% Wilson intervals. Row numbers follow Tab.~2.}
\label{tab:suppl_confusion}
\end{table*}

\subsection{Remaining PROVE Errors}
\label{sec:suppl_errors}

Of PROVE's 206 errors, 160 are wrong acceptances and 46 are success rejections (\cref{tab:suppl_confusion}).
PROVE therefore errs mainly by accepting failures, which matches the hard cases in the main paper: transient activation and millimeter-scale closure gaps that look complete in terminal frames.
PROVE-Lite's 572 errors split the other way, with 170 wrong acceptances and 402 success rejections.
Without the initial scene and the goal image, it rejects more valid outcomes.
For a finer breakdown, we group ten actions into appliance activation
(TurnOnElectricKettle and TurnOnMicrowave),
fine closure (CloseFridge, CloseDishwasher,
and CloseBlenderLid),
placement (CoffeeSetupMug,
PickPlaceCounterToCabinet,
and PickPlaceSinkToCounter),
and other actions (SlideDishwasherRack
and TurnSinkSpout).
Of PROVE's 206 errors over all 65 actions, 157 (76.21\%) fall in the appliance-activation and fine-closure groups.
Within these ten actions, PROVE-Lite makes 88 errors, of which 72 (81.82\%) fall in the fine-closure and placement groups.

\subsection{Diagnosis and Head Consistency}
\label{sec:suppl_diagnostic_metrics}

Against PROVE-derived labels, status accuracy is 87.80\%
(0.817 macro-F1), class accuracy is 80.77\%
(0.542 macro-F1), and the MAE of $C$ is 0.165.
Among rollouts predicted successful, 10.8\% have $C<1$;
this subset has a mean reward of 0.877.

The condition-level metrics are evaluated against
PROVE-derived labels.
Their downstream utility is assessed by comparing binary
and graded rewards from the same frozen evaluator.
Graded achieves higher success under both PROVE and
native-predicate evaluation
(\cref{sec:suppl_ppo_results}).

\Cref{tab:suppl_threshold_sensitivity} varies the success
threshold of the DecisionHead.
For thresholds from 0.3 to 0.6, agreement stays within
1.2 percentage points of the result at the default
threshold of 0.5.
Lower thresholds reduce success rejection while
increasing wrong acceptance.

\begin{table}[!htb]
\centering
\small
\begin{tabular}{cccc}
\toprule
\textbf{Threshold} & \textbf{Agree.} & \textbf{Wrong acc.} & \textbf{Succ.\ rej.} \\
\midrule
0.3 & 90.18 & 14.97 & 6.06 \\
0.4 & 91.15 & 8.14 & 9.36 \\
0.5 & 91.20 & 6.21 & 10.69 \\
0.6 & 90.02 & 5.44 & 13.29 \\
0.7 & 88.37 & 4.38 & 16.91 \\
\bottomrule
\end{tabular}
\caption{PROVE-Lite on the benchmark under different success thresholds (\%).}
\label{tab:suppl_threshold_sensitivity}
\end{table}

\subsection{Timing Protocol}
\label{sec:suppl_efficiency}

Latency and memory in Tab.~4 are measured on one NVIDIA RTX PRO 4500 GPU.
We timed 200 rollouts per evaluator, processing one rollout at a time.
Timing includes image loading and preprocessing.
For PROVE, the timer covers condition generation and terminal
verification using precomputed goal images. Goal-image generation is
outside the timed interval.
The PROVE timing script uses 10 separate warm-up rollouts.

\subsection{Decomposing Policy Disagreements}
\label{sec:suppl_reeval_decomp}

\Cref{tab:suppl_reeval_decomp} decomposes disagreements
into native-only successes, where native predicates
accept a rollout and PROVE rejects it, and PROVE-only
successes, where the reverse holds.
Native-only successes outnumber PROVE-only successes
for every policy, explaining the higher success rates
reported by native predicates.

The decomposition also clarifies the ranking reversal
between PRTS and RLDX-1.
Although native predicates accept more RLDX-1 rollouts,
PROVE rejects a similar number of native-only successes
for the two policies (2,039 for PRTS and 2,092 for RLDX-1)
while accepting substantially more PROVE-only successes
for PRTS (598 versus 150).
Consequently, PROVE accepts 2,854 PRTS rollouts and
2,607 RLDX-1 rollouts.
GR00T N1.5 and RLDX-1 differ by 66 accepted rollouts
under PROVE, corresponding to 1.02 percentage points.

Across all 8,677 disagreements, blind human review
supports PROVE in 8,280 cases (95.42\%;
95\% Wilson interval: 94.96--95.84\%).
The support rate ranges from 94.77\% to 95.87\%
across the four policies.

\begin{table*}[t]
\centering
\small
\setlength{\tabcolsep}{0pt}
\begin{tabular*}{\textwidth}{@{\extracolsep{\fill}}lrrrrrcc@{}}
\toprule
\textbf{Policy} & \textbf{Native S} & \textbf{PROVE S} & \textbf{Native-only S} & \textbf{PROVE-only S} & \textbf{Disagree} & \textbf{N / P} & \textbf{Side with PROVE (95\% CI)} \\
\midrule
GR00T N1.5 & 3,449 & 2,673 & 809 & 33 & 842 & 44 / 798 & 94.77 (93.06--96.08) \\
RLDX-1 & 4,549 & 2,607 & 2,092 & 150 & 2,242 & 117 / 2,125 & 94.78 (93.78--95.63) \\
PRTS & 4,295 & 2,854 & 2,039 & 598 & 2,637 & 109 / 2,528 & 95.87 (95.04--96.56) \\
Awomo-0.5I & 5,639 & 3,009 & 2,793 & 163 & 2,956 & 127 / 2,829 & 95.70 (94.91--96.38) \\
\midrule
All & 17,932 & 11,143 & 7,733 & 944 & 8,677 & 397 / 8,280 & 95.42 (94.96--95.84) \\
\bottomrule
\end{tabular*}
\caption{Policy disagreements decomposed. Success counts are out of 6,500 rollouts per policy and follow from Tab.~3. Native-only S: native accepts, PROVE rejects; PROVE-only S: the reverse.}
\label{tab:suppl_reeval_decomp}
\end{table*}

\begin{table}[t]
\centering
\footnotesize
\setlength{\tabcolsep}{3pt}
\begin{tabular*}{\columnwidth}{@{\extracolsep{\fill}}lcccc@{}}
\toprule
& \multicolumn{2}{c}{\textbf{PROVE-judged}} & \multicolumn{2}{c}{\textbf{Native-predicate}} \\
\cmidrule(lr){2-3}\cmidrule(l){4-5}
\textbf{Group} & \textbf{Succ.\ (\%)} & \textbf{$\Delta$} & \textbf{Succ.\ (\%)} & \textbf{$\Delta$} \\
\midrule
Base policy & $50.17\pm0.77$ & -- & $50.58\pm0.88$ & -- \\
Simulator & $52.00\pm0.81$ & $+1.83$ & $52.67\pm0.90$ & $+2.09$ \\
Binary & $54.58\pm0.88$ & $+4.41$ & $54.08\pm1.03$ & $+3.50$ \\
Graded & $\mathbf{56.58\pm0.91}$ & $\mathbf{+6.41}$ & $\mathbf{56.83\pm1.45}$ & $\mathbf{+6.25}$ \\
\bottomrule
\end{tabular*}
\caption{PPO results on 1,200 held-out episodes per group (mean $\pm$ s.d.\ across four seeds).
$\Delta$: gain over the base policy in percentage points.}
\label{tab:suppl_ppo_results}
\end{table}


\section{PPO Protocol}
\label{sec:suppl_ppo}

\subsection{Groups and Data}
\label{sec:suppl_ppo_groups}

All groups start from \texttt{RLinf-pi0-robocasa-\allowbreak pretrain-human300} (\cref{tab:suppl_ppo_groups}).
The PPO benchmark covers 15 actions and four seeds (42--45).
For each action, rollouts 0--7 in 10 scenes form 80 training initializations, and rollouts 8--9 form 20 test initializations.
These 20 test initializations are fixed by rollout index and shared across all groups and seeds.
Each trained policy collects eight trajectories per update over five updates, so 40 training trajectories per action, group, and seed. Each group is tested on $15\times4\times20=1{,}200$ episodes.
The reward is given only at termination, and GAE propagates it across action chunks.
The Binary group uses $R=S$, i.e., $\lambda=1$; the Graded group uses $R=0.8S+0.2C$, i.e., $\lambda=0.8$.
Both groups use predictions from the same frozen PROVE-Lite: Binary uses the success prediction alone, whereas Graded combines it with the condition-based completion score.

\begin{table}[!htb]
\centering
\small
\begin{tabular}{lll}
\toprule
\textbf{Group} & \textbf{Update} & \textbf{Terminal reward} \\
\midrule
Base policy & None & -- \\
Simulator & PPO, 5 iterations & Native predicate, $\{0,1\}$ \\
Binary & PPO, 5 iterations & $R=S$ \\
Graded & PPO, 5 iterations & $R=\lambda S+(1-\lambda)C$ \\
\bottomrule
\end{tabular}
\caption{PPO groups. Binary uses PROVE-Lite's outcome $S$.
Graded uses $S$ and completion score $C$, with $\lambda=0.8$.}
\label{tab:suppl_ppo_groups}
\end{table}

\subsection{Hyperparameters and Hardware}
\label{sec:suppl_ppo_hparams}

\begin{table}[!htb]
\centering
\small
\begin{tabular}{ll}
\toprule
\textbf{Parameter} & \textbf{Value} \\
\midrule
Optimizer & AdamW \\
Learning rate & Policy $5\times10^{-6}$; value $1\times10^{-4}$ \\
Discount $\gamma$ & 0.99 \\
GAE $\lambda$ & 0.95 \\
Clip range & 0.2 \\
Value clip & 0.2 \\
Entropy coefficient & 0 \\
Global batch size & 8 \\
PPO epochs per update & 1 \\
Gradient-norm clipping & 1.0 \\
Policy LoRA & $r=16$, all linear layers \\
Parallel environments & 8 \\
Action chunks per step & 5 \\
Adam $(\beta_1,\beta_2)$ & (0.9, 0.95) \\
Weight decay & 0.01 \\
Episode horizon & 300--900 steps per action \\
\bottomrule
\end{tabular}
\caption{PPO hyperparameters.}
\label{tab:suppl_ppo_hparams}
\end{table}

The three PPO training groups run concurrently on
three separate NVIDIA H100 GPUs, one per reward group.
A frozen PROVE-Lite model is deployed on a fourth
NVIDIA H100 GPU and serves reward requests asynchronously.
Failed reward requests or invalid reward outputs
trigger retries.

\subsection{Aggregate Results Across Seeds}
\label{sec:suppl_ppo_results}

\Cref{tab:suppl_ppo_results} expands Table~6 of the main paper.
The graded reward also ranks first under native predicates.
Under native-predicate evaluation, Graded exceeds Binary
by 2.75 percentage points and Simulator by 4.17 percentage
points.

\paragraph{Evaluation coverage}
We evaluate each group on 15 actions with 20 episodes
per action under each of four random seeds (42--45),
giving 1,200 episodes per group.

\begin{figure}[!htb]
\centering
\includegraphics[width=\linewidth]
{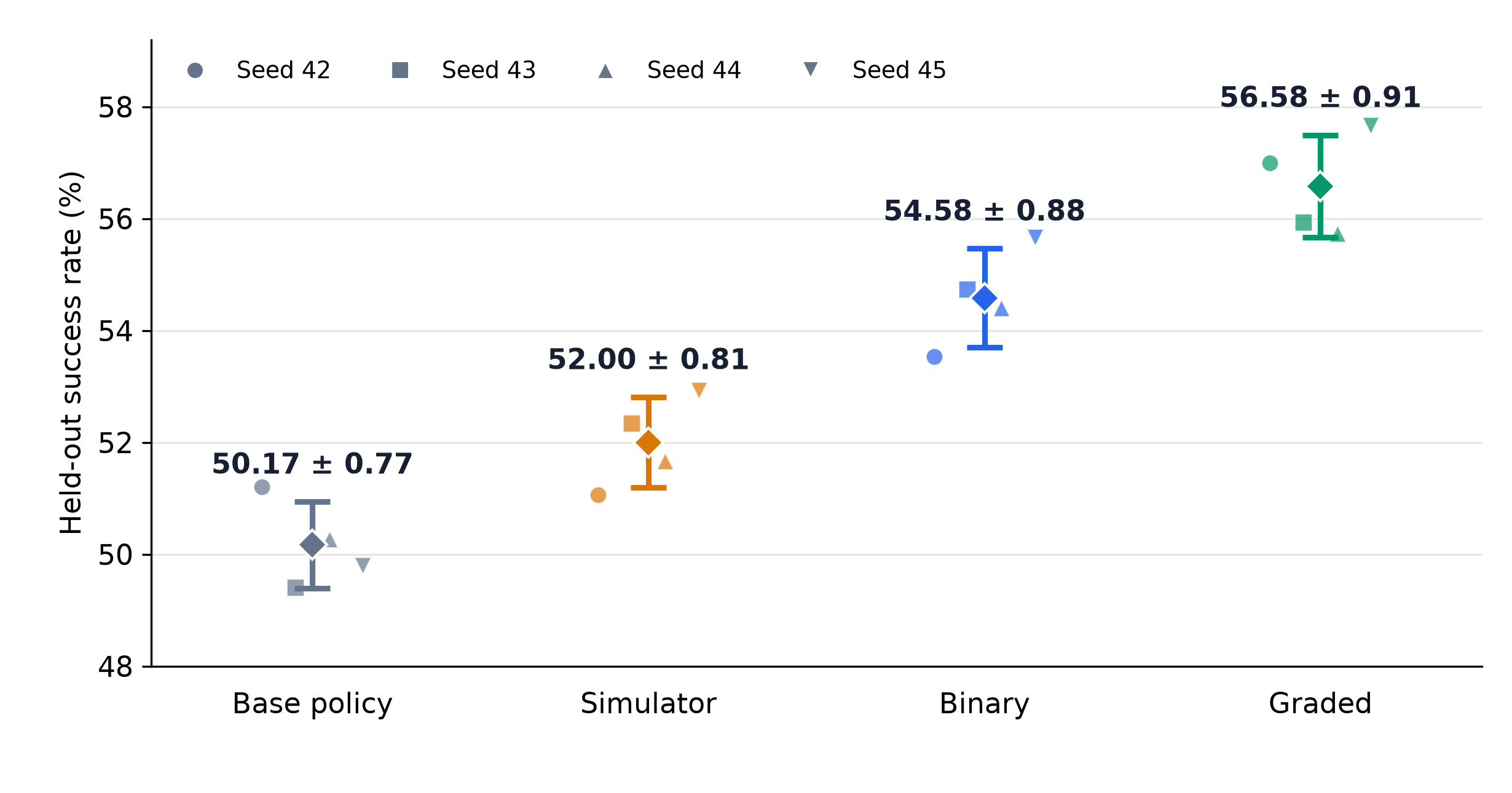}
\caption{\textbf{PPO results.}
Final held-out success rates across four seeds
(42--45).
Small markers show individual seeds;
diamonds and error bars indicate the mean
and sample standard deviation.
}
\label{fig:suppl_ppo_results}
\end{figure}

\twocolumn[{%
\section{Qualitative Examples}
\label{sec:suppl_cases}
\begin{minipage}{\textwidth}
\centering
\small
\textbf{Instruction.} Turn on the toaster oven by setting the timer.\par\medskip
\begin{minipage}[t]{0.32\textwidth}
\centering
\includegraphics[width=\linewidth,height=0.105\textheight,keepaspectratio]{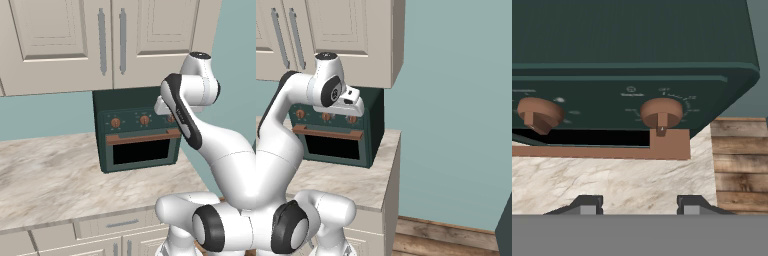}\par
Initial scene
\end{minipage}\hfill
\begin{minipage}[t]{0.32\textwidth}
\centering
\includegraphics[width=\linewidth,height=0.105\textheight,keepaspectratio]{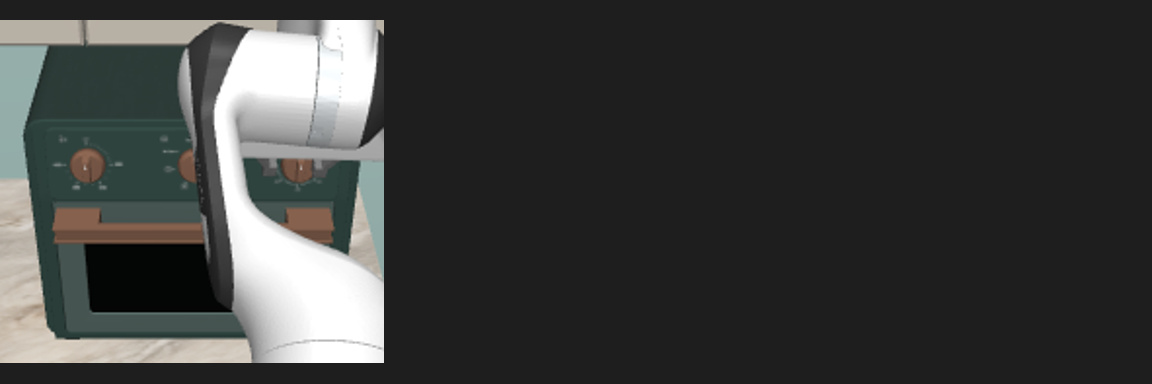}\par
Target-object reference
\end{minipage}\hfill
\begin{minipage}[t]{0.32\textwidth}
\centering
\includegraphics[width=\linewidth,height=0.105\textheight,keepaspectratio]{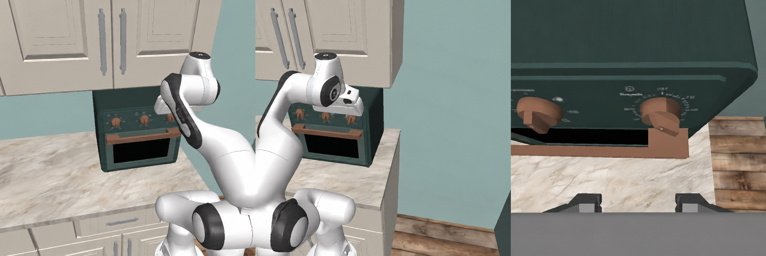}\par
Generated goal image
\end{minipage}\par\medskip
\begin{minipage}[t]{0.32\textwidth}
\raggedright
\textbf{M1 ($C_1$) --- Instruction only}\par\smallskip
Toaster oven timer is set to a non-zero value\par
Toaster oven power indicator light is illuminated\par
Toaster oven control knob or button is in the `on' position\par
\textbf{Toaster oven door is closed} \emph{(removed in M2)}
\end{minipage}\hfill
\begin{minipage}[t]{0.32\textwidth}
\raggedright
\textbf{M2 ($C_2$) --- Scene and object}\par\smallskip
Toaster oven timer is set to a non-zero value\par
\textbf{Toaster oven power indicator light is illuminated} \emph{(removed in M3)}\par
Toaster oven control knob or button is in the `on' position
\end{minipage}\hfill
\begin{minipage}[t]{0.32\textwidth}
\raggedright
\textbf{M3 ($C_3$) --- Goal-guided refinement}\par\smallskip
Toaster oven timer is set to a non-zero value\par
Toaster oven control knob or button is in the `on' position
\end{minipage}
\captionof{figure}{\textbf{Progressive specification for turning on a toaster oven.}
M2 drops the closed-door requirement after incorporating the initial scene and target-object reference.
M3 drops the power-indicator requirement after incorporating the generated goal image, retaining the timer and control-position conditions.
Bold text marks the conditions removed at the next stage.}
\label{fig:suppl_m1m3_case}
\end{minipage}

\vspace{1.2em}
\begin{minipage}{\textwidth}
\centering
\begin{minipage}[t]{0.40\textwidth}
\centering
\includegraphics[width=\linewidth]{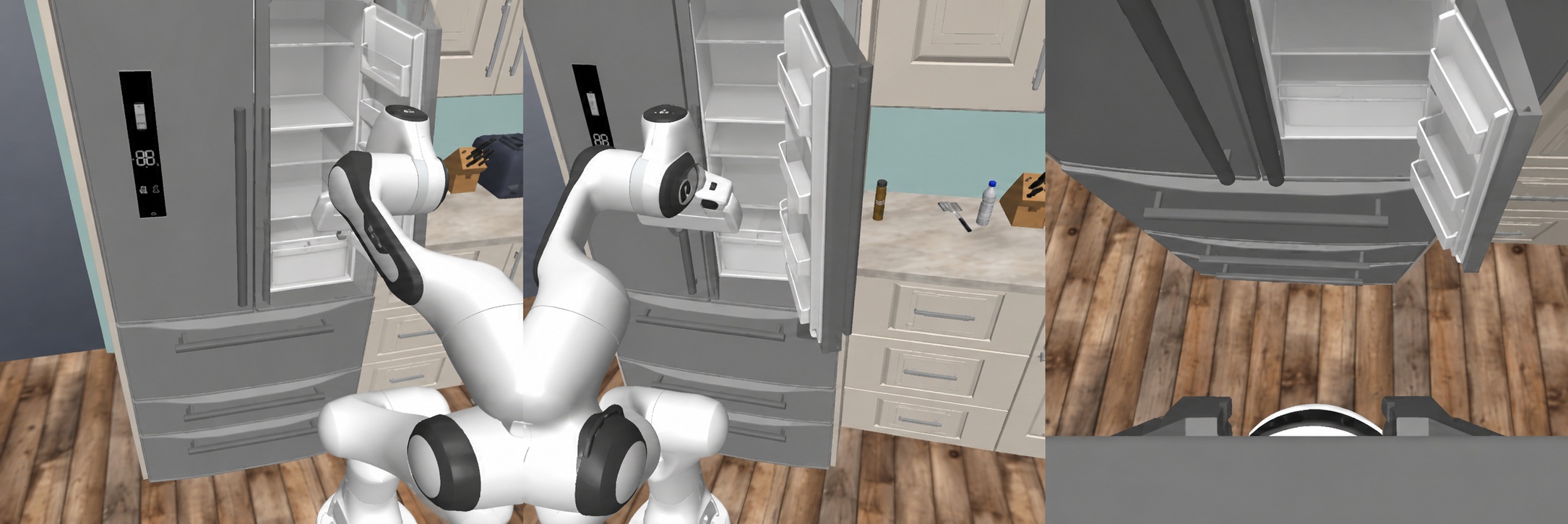}\par
\smallskip
{\small (a) Open the fridge doors.}
\end{minipage}\hspace{0.04\textwidth}%
\begin{minipage}[t]{0.40\textwidth}
\centering
\includegraphics[width=\linewidth]{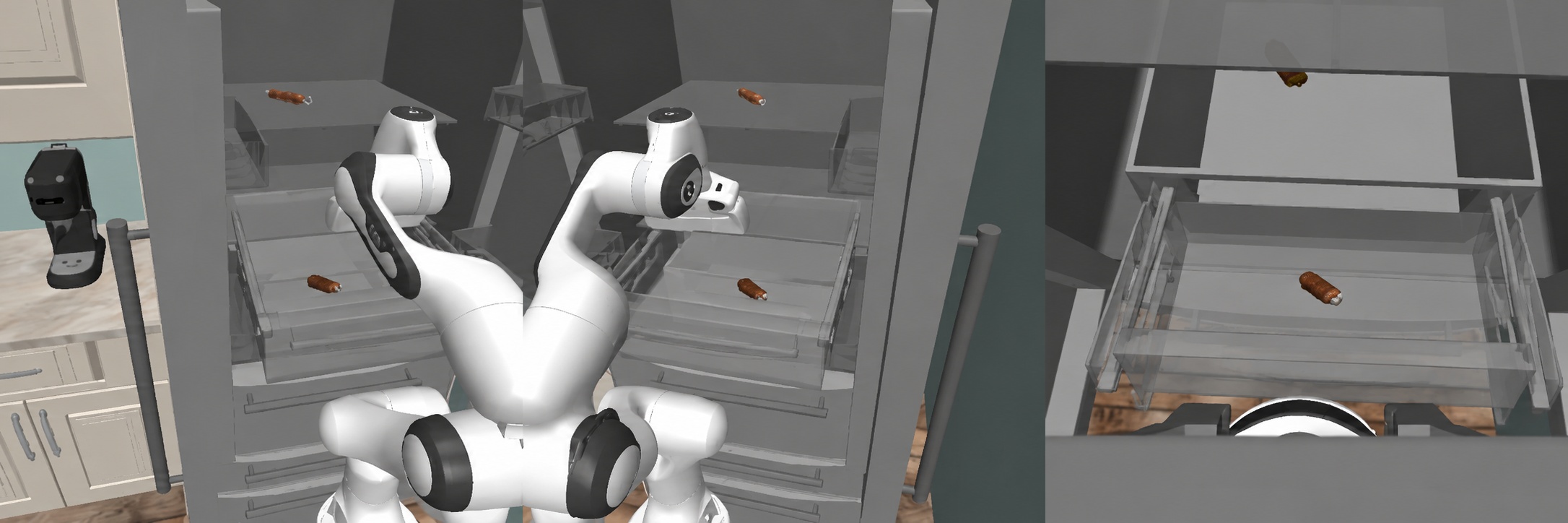}\par
\smallskip
{\small (b) Fully open the fridge drawer.}
\end{minipage}
\captionof{figure}{\textbf{Artifacts in generated goal images.}
The generation review flagged (a) a duplicate handle at the door edge
and (b) a lower food item replaced by a duplicate of the elongated item above.
PROVE uses generated goal images to refine success conditions and
verifies those conditions using recorded terminal observations.}
\label{fig:suppl_goal_artifacts}
\end{minipage}

\vspace{1.2em}
\begin{minipage}{\textwidth}
\centering
\begin{minipage}[t]{0.49\textwidth}
\small
\raggedright
\textbf{(a) CoffeeSetupMug}
\hfill {\footnotesize\texttt{t00-s01-r04}}\par
\smallskip
{\footnotesize Human: \textbf{FAILURE}
\enspace PROVE: \textbf{FAILURE}
\enspace PROVE-Lite: \textbf{SUCCESS}\par}
\smallskip
\includegraphics[width=\linewidth]{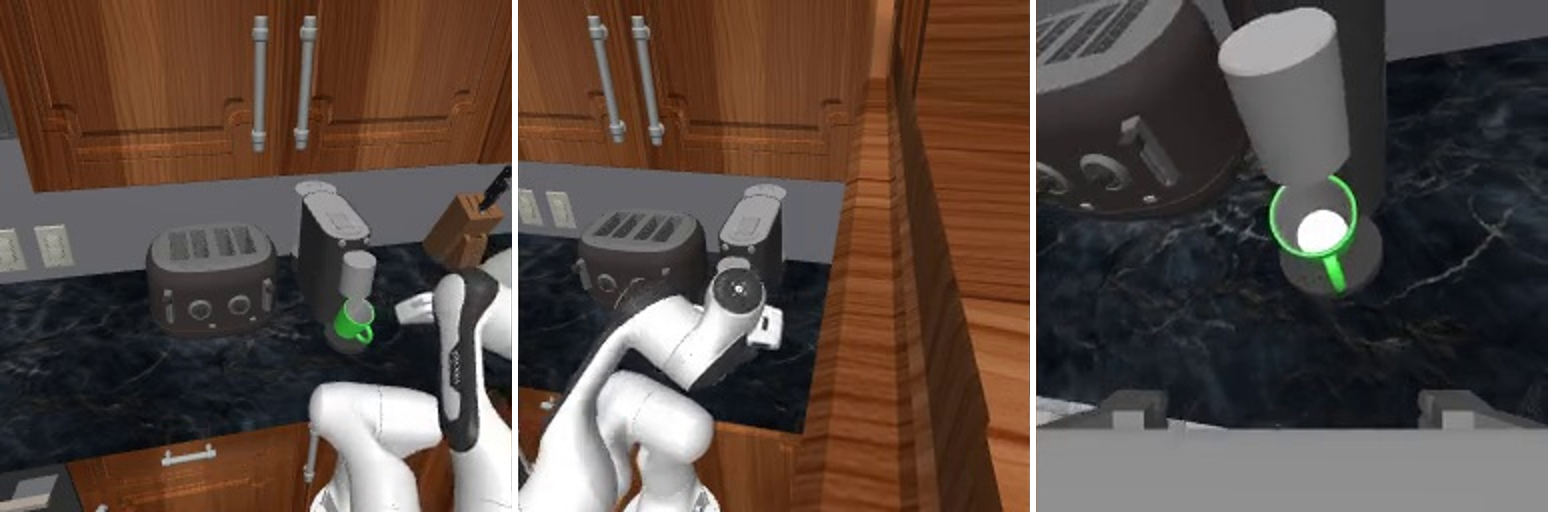}
\par\smallskip
\textbf{PROVE.}
The mug is not resting on the coffee-machine tray.\par
\smallskip
\textbf{PROVE-Lite.}
\texttt{mug\_aligned}: SATISFIED / \texttt{aligned};
\texttt{stable\_placement}: SATISFIED / \texttt{supported}.\par
\smallskip
\textbf{Evaluation error.}
PROVE-Lite falsely accepts the rollout.
\end{minipage}\hfill
\begin{minipage}[t]{0.49\textwidth}
\small
\raggedright
\textbf{(b) PickPlaceSinkToCounter}
\hfill {\footnotesize\texttt{t05-s05-r07}}\par
\smallskip
{\footnotesize Human: \textbf{SUCCESS}
\enspace PROVE: \textbf{FAILURE}
\enspace PROVE-Lite: \textbf{SUCCESS}\par}
\smallskip
\includegraphics[width=\linewidth]{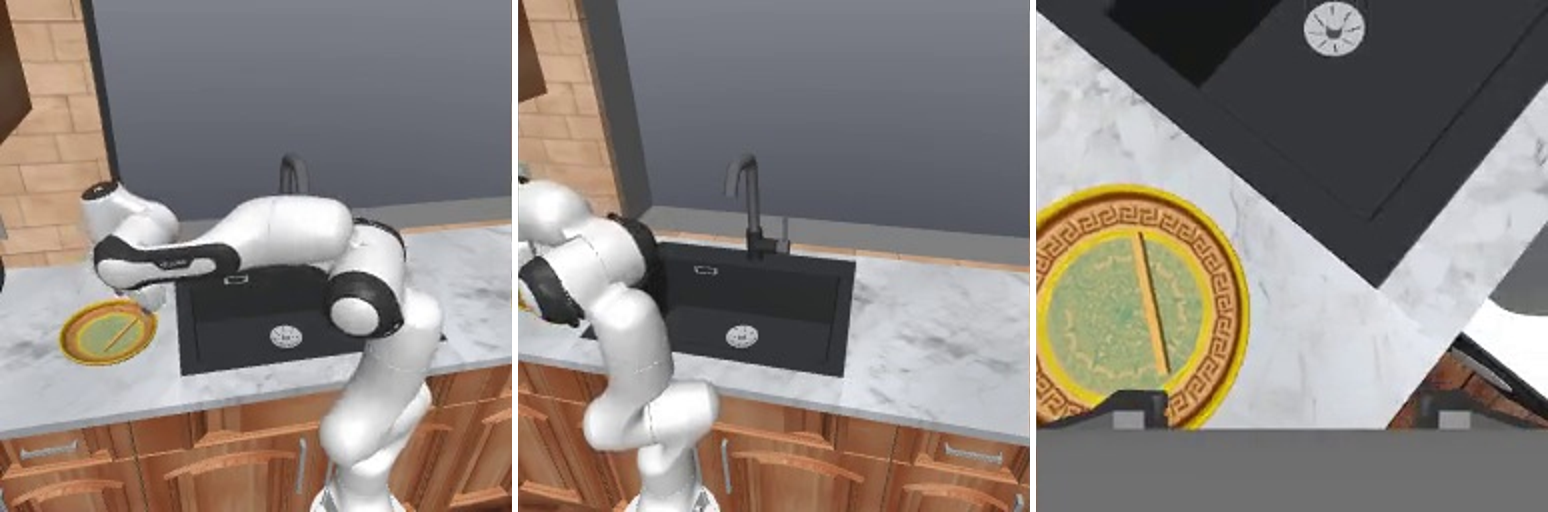}
\par\smallskip
\textbf{PROVE.}
The cheese extends beyond the plate; its explanation also
questions the plate location.\par
\smallskip
\textbf{PROVE-Lite.}
\texttt{object\_counter}: SATISFIED / \texttt{supported};
\texttt{stable\_placement}: SATISFIED / \texttt{supported}.\par
\smallskip
\textbf{Evaluation error.}
PROVE falsely rejects a human-labeled success;
PROVE-Lite agrees with the human.
\end{minipage}
\captionof{figure}{\textbf{Failure cases of PROVE and PROVE-Lite.}
Each case shows the final left, right, and wrist views,
the human verdict, and the evaluator predictions.
(a) PROVE-Lite incorrectly accepts a failed mug-placement rollout.
(b) PROVE incorrectly rejects a successful placement rollout.
In the placement case in Figure~4 of the main paper,
both PROVE and PROVE-Lite incorrectly reject a valid
outcome. Here, PROVE-Lite correctly accepts the rollout
despite PROVE's false rejection.
The explanations and condition outputs are model predictions.}
\label{fig:suppl_failure_cases}
\end{minipage}
\vspace{0.5em}
}]
\end{document}